\RequirePackage[svgnames]{xcolor}

\documentclass[11pt,a4paper]{mystyle}

\usepackage[all]{hypcap}
\usepackage[svgnames]{xcolor}
\usepackage[comma,authoryear,compress]{natbib}
\usepackage{amsmath,amsfonts}
\usepackage{algorithmic}
\usepackage{algorithm}
\usepackage{array}
\usepackage{textcomp}
\usepackage{stfloats}
\usepackage{url}
\usepackage{verbatim}
\usepackage{graphicx}
\usepackage{xcolor}
\usepackage[font=small,labelfont=bf]{caption}
\usepackage{subcaption}
\usepackage{textcomp}
\usepackage{stfloats}
\usepackage{amssymb}
\usepackage{latexsym}
\usepackage[utf8]{inputenc}
\usepackage{microtype}
\usepackage{inconsolata}
\usepackage{ragged2e}
\usepackage{balance}
\usepackage{multirow}
\usepackage{multicol}
\usepackage{makecell}
\usepackage{tabularx,booktabs}
\usepackage{enumitem}
\usepackage{xspace}
\usepackage{fontawesome5}
\usepackage{float}
\usepackage{pifont}
\usepackage{microtype}
\usepackage{ulem}
\usepackage{bm}
\usepackage{autonum}
\usepackage{colortbl}
\usepackage{longtable}
\usepackage{hyperref}
\usepackage{wrapfig}
\usepackage{tcolorbox}
\usepackage{floatflt}
\usepackage{arydshln}
\usepackage{lineno}

\usepackage{dsfont}
\usepackage{cleveref}
\usepackage{bxcoloremoji}

\definecolor{bettergreen}{rgb}{0.13, 0.55, 0.13}
\definecolor{verylightgray}{gray}{0.94}
\definecolor{redv}{HTML}{C00000}
\definecolor{orangev}{HTML}{ce7a1e}
\definecolor{yellowv}{HTML}{d6a03d} 
\definecolor{bluev}{HTML}{0070C0}
\definecolor{grayv}{HTML}{707070}

\newcommand{\FrameName}{{\scshape \kern-0.5pt N\kern-0.5pt i\kern-0.5pt n\kern-0.5pt t}\xspace}
\newcommand{\DatasetName}{{\scshape \kern-0.5pt N\kern-0.5pt e\kern-0.5pt w\kern-0.5pt s\kern-0.5pt I\kern-0.5pt n\kern-0.5pt t}\xspace}
\newcommand{\ModelName}{{\scshape \kern-0.5pt D\kern-0.5pt M\kern-0.5pt I\kern-0.5pt n\kern-0.5pt t}\xspace}
\newcommand{\LLMName}{{\scshape \kern-0.5pt D\kern-0.5pt M\kern-0.5pt G}\xspace}

\newtheorem{myDef}{Definition}
\newcommand{\redxmark}{\textcolor{red}{\ding{53}}}

\definecolor{myblue}{HTML}{3889fe}
\definecolor{myorange}{HTML}{b2561a}
\definecolor{mygreen}{HTML}{6ea13f}
\definecolor{lightgrayv}{HTML}{F4F3F8} 
\definecolor{lightbluev}{HTML}{F5F9FD} 
\definecolor{grayv}{HTML}{707070}

\title{Exploring news intent and its application:\\A theory-driven approach}

\runningtitle{Exploring news intent and its application: A theory-driven approach}

\author[1,2]{\href{https://scholar.google.com/citations?user=fMCBAt4AAAAJ}{\textcolor{black}{Zhengjia Wang}}}
\author[1]{\href{https://scholar.google.com/citations?user=hGZwK0cAAAAJ}{\textcolor{black}{Danding Wang}}}
\author[1]{\href{https://sheng-qiang.github.io/}{\textcolor{black}{Qiang Sheng}}}
\author[1,2]{\href{https://scholar.google.com/citations?user=fSBdNg0AAAAJ}{\textcolor{black}{Juan Cao}}}
\author[3]
{\href{mailto:siyuanma@um.edu.mo}
{\textcolor{black}{Siyuan Ma}}}
\author[4]
{\href{mailto:haonancheng@cuc.edu.cn}
{\textcolor{black}{Haonan Cheng}}}

\affil[1]{Media Synthesis and Forensics Lab, Institute of Computing Technology, Chinese Academy of Sciences}
\affil[2]{University of Chinese Academy of Sciences}
\affil[3]{University of Macau}
\affil[4]{Communication University of China}

\correspondingauthor{Juan Cao}
\emails{
    {\{\href{mailto:wangzhengjia21b@ict.ac.cn}{wangzhengjia21b},\href{mailto:wangdanding@ict.ac.cn}{wangdanding},\href{mailto:shengqiang18z@ict.ac.cn}{shengqiang18z},\href{mailto:caojuan@ict.ac.cn}{caojuan}\}@ict.ac.cn},
    {\href{siyuanma@um.edu.mo}{siyuanma@um.edu.mo},}
    {\href{haonancheng@cuc.edu.cn}{haonancheng@cuc.edu.cn}
    }
}

\begin{document}

\begin{abstract}
Understanding the intent behind information is crucial. However, news as a medium of public discourse still lacks a structured investigation of perceived news intent and its application. To advance this field, this paper reviews interdisciplinary studies on intentional action and introduces a conceptual deconstruction-based news intent understanding framework (\FrameName). This framework identifies the components of intent, facilitating a structured representation of news intent and its applications. Building upon \FrameName, we contribute a new intent perception dataset. Moreover, we investigate the potential of intent assistance on news-related tasks, such as significant improvement (+$2.2\%$ macF1) in the task of fake news detection. We hope that our findings will provide valuable insights into action-based intent cognition and computational social science.
\vspace{15pt}

\coloremojicode{1F4C5} \textbf{Date}: June 17, 2025

\coloremojicode{1F3E0} \textbf{Project}: \href{https://github.com/ICTMCG/NewsInt}{https://github.com/ICTMCG/NewsInt}

\coloremojicode{1F4AC} \textbf{Journal}: Information Processing \& Management

\textcolor{myblue}{\faLink} \textbf{DOI}: https://doi.org/10.1016/j.ipm.2025.104229
\end{abstract}

\maketitle

\vspace{25pt}

\begin{figure}[H]
    \centering
    \includegraphics[width=.85\linewidth]{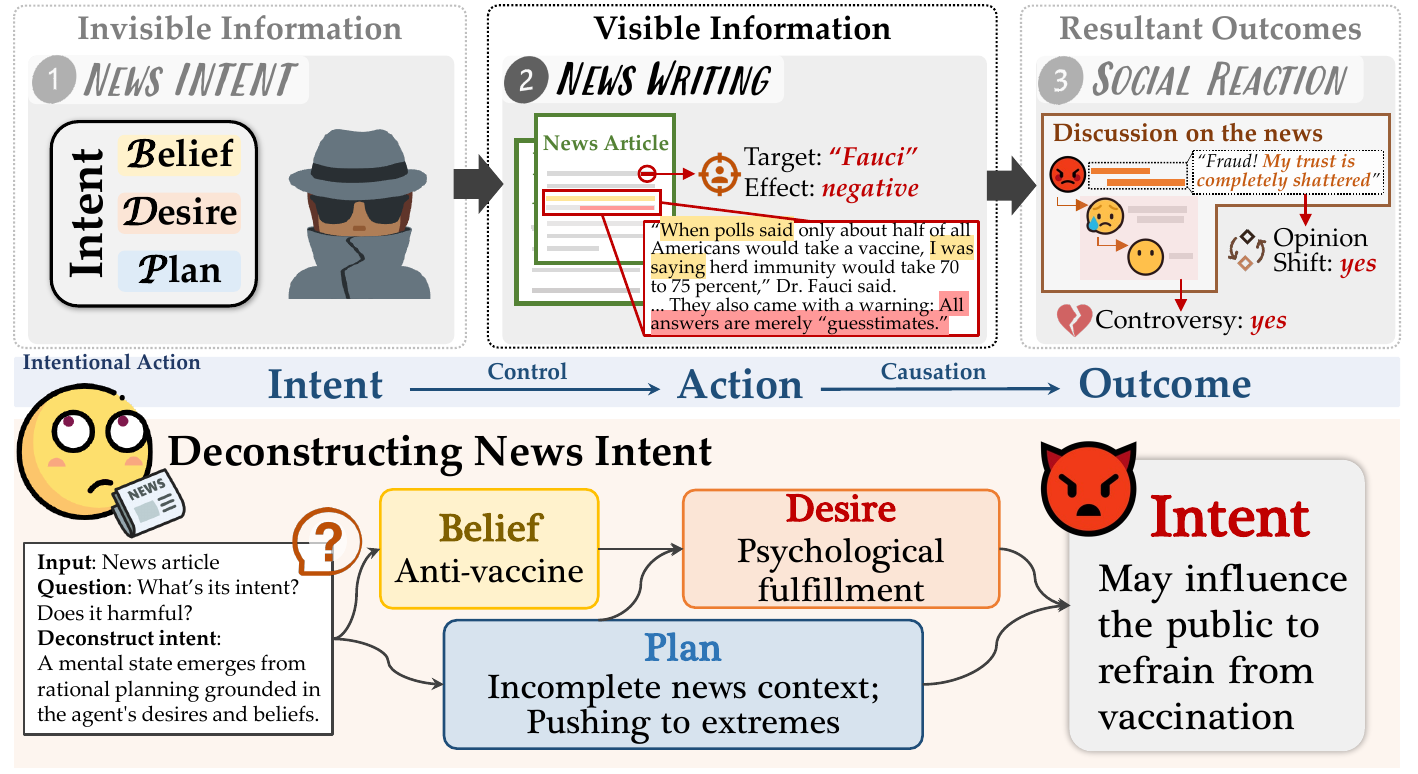}
    \caption{Process of news production. News intent controls news writing as a rational action, subsequently eliciting social reactions as outcomes. This motivates us to deconstruct news intent from aspects of rational action and outcome.}
    \label{fig:motive}
\end{figure}

\section{Introduction} \label{sec:intro}

\begin{center}
\begin{quote}
\textit{``The most important thing in communication is hearing what isn't said.''} \\
\hfill --- Peter Drucker
\end{quote}
\end{center}

Intent refers to the purpose or goal behind an action~\citep{bratman1984two}, which involves the mental state of aiming to achieve a specific outcome~\citep{bratman2009intention}. 
In various aspects, such as law, psychology, and artificial intelligence, understanding intent helps interpret actions and predict future behavior accurately~\citep{anscombe2000intention}.
Recent research has focused on intent in interpersonal communication~\citep{mittal2024towards, zhang2024mintrec}, human-machine communication~\citep{weld2022survey}, individual consumption~\citep{chen2022intent}, and pragmatic language~\citep{yerukola2024pope}. Similarly, news as a medium of public discourse, identifying the intent behind news has garnered increasing interest among researchers.
However, existing research on news intent struggles with the intricate manifestations of this abstract concept in news, often resorting to unstructured descriptions to analyze news intent \citep{gabriel2022misinfo}, lacking standardization for effective quantitative analysis and comparison. Limited structured research on news intent focuses on specific cases, such as detecting advertising intent in health news \citep{kang2019health}, relying on particular subjects rather than broader theoretical frameworks, which restricts their general applicability. 

To bridge this gap, our research is guided by the following research objectives (ROs):
\begin{itemize}[itemsep=2pt,topsep=2pt,parsep=0pt,leftmargin=44pt]
    \item[\textbf{RO1:}] To establish a foundation for news intent research from a theory-driven approach and to explore what news intent is and the benefits of deconstructing it.
    \item[\textbf{RO2:}] To develop a comprehensive and rich dataset for news intent perception and investigate how the model perceives news intent and how our framework facilitates this perception.
    \item[\textbf{RO3:}] To examine and leverage the implications of news intent and to explore how perceived news intent can be utilized to deepen news understanding and achieve practical value.
\end{itemize}

To achieve these objectives, we draw insights from multidisciplinary theories and propose a deconstruction-based framework to facilitate the perception of news intent, resulting in a structured and more universally applicable framework. By modeling news production as an intentional action \citep{cushman2008intentional, anscombe2000intention, bratman2009intention}, this approach provides a comprehensive understanding of how news intent shapes news production and causes specific social impacts (\figureautorefname~\ref{fig:motive}). Motivated by this observation, we present the first \textbf{Conceptual Deconstruction-based} \textbf{\uline{N}ews \uline{INT}ent Understanding framework} (\textbf{\FrameName}), deconstructing news intent from perspectives of rational action \citep{gergely2003teleological} and outcomes \citep{malle1997folk, cushman2015deconstructing}. Specifically, \FrameName deconstructs news intent into three key components: belief, desire, and plan, each representing a core aspect of intent (detailed in Section~\ref{sec:framework}). Our emphasis on both internal components and external connections of news intent marks a significant improvement from previous research, which overlooks the interrelations between these elements.

Based on \FrameName, we curate a news intent perception dataset, \textbf{\DatasetName}, with fine-grained labels of belief, desire, plan, and intent polarity, aligned with human perception following a two-stage annotation process. This dataset not only surpasses existing news datasets in its detailed annotation of news intent but also facilitates a systematic study of news production as an intentional action: it covers multiple topics and diverse news types, capturing substantial social reactions and effectively simulating real-world intent perception evaluation.
Moreover, surprisingly, we find that intent can aid in revealing reporting motivations, predicting potential impacts, and enhancing resilience to misinformation and persuasive content in general news. We design methods and empirically demonstrate the implications of news intent in important news-related tasks, such as fake news detection, news popularity prediction, and propaganda detection. Our findings indicate that intent perception can enhance news understanding, leading to performance improvements across various application scenarios, such as an average relative improvement of 2.8\% in F1$_{\text{fake}}$ for fake news detection, underscoring the practical value of news intent.

Our main contributions are:
\begin{itemize}[itemsep=2pt,topsep=2pt,parsep=0pt,leftmargin=18pt]
    \item \textbf{News intent deconstruction framework.} We perform a systematic investigation of interdisciplinary studies and propose the first conceptual deconstruction framework, achieving the structured news intent representation.
    \item \textbf{News intent perception dataset.}
    We build a new dataset \DatasetName, dedicated to news intent perception, comprising 12,959 news articles with 142.5k fine-grained labels\footnote{Relevant resources will be made publicly available.}.
    \item \textbf{News intent application.}
    We propose methods to enhance the model's perception of news intent, with experiments on three application tasks demonstrating its practical value in improving news understanding.
\end{itemize}

\noindent\textbf{Organization.}
The rest of this paper is organized as follows. Section 2 introduces the theoretical foundation of our proposed framework (RO1). Section 3 provides details of dataset curation (RO2). The proposed methods for identifying news intent (RO2) and enhancing news applications (RO3) are presented in Section 4. In Section 5, we conduct experiments and report the results of our methods in news intent identification, as well as explore the practical value of news intent through experiments on three application tasks. Section 6 discusses the theoretical and practical implications of this work. Section 7 reviews the related works of this research. Concluding remarks are provided in Section 8.

\section{News Intent Deconstruction} \label{sec:framework}
\begin{figure*}[t]
    \centering
    \includegraphics[width=.98\linewidth]{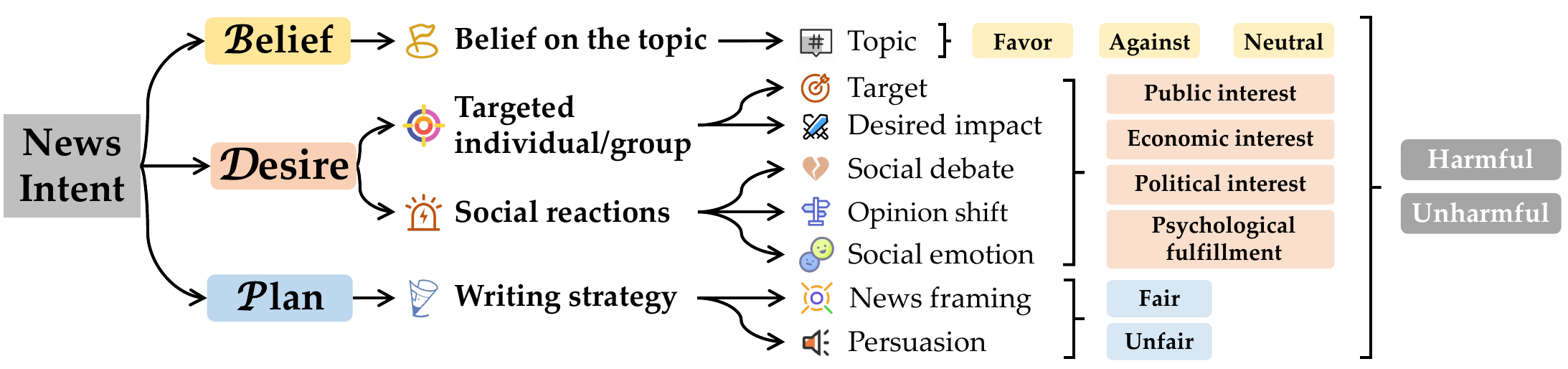}
    \caption{The proposed \FrameName framework. \FrameName deconstructs the concept of news intent into beliefs, desires, and plans using interdisciplinary theories. It further situates these elements within the specific context of news to investigate the concrete manifestations of news intent. \label{fig:frame}}
\end{figure*}

Unlike previous studies that directly handle intent from language descriptions \citep{gabriel2022misinfo}, we start from the conceptual deconstruction of news intent to clarify intent ambiguity, step by step.
Then, inspired by intentional action studies \citep{anscombe2000intention, cushman2008intentional, bratman2009intention}, we delve into the process of news generation to instantiate the concept of intent and its elements from news writing and social reactions.
The proposed framework is illustrated in \figurename~\ref{fig:frame}.

\subsection{Conceptual Deconstruction}\label{ssec:deconstruct}

This section aims to facilitate a structured analysis of intent and its deconstruction from the concept perspective, as explored by psychologists, philosophers, and cognitive scientists over several decades.

\noindent \textbf{Intent.}
People do things intentionally, and people intend to do things \citep{bratman1984two}. The concept of ``intentional action'' \citep{ cushman2008intentional} requires understanding the intent behind deliberate behavior, linking human intent closely with actions. Based on psychological research, we offer a deconstructed theory-driven definition of intent: 
\begin{myDef}[Intent]
\label{def:intent}
Intent refers to a cognitive state that emerges from rational planning, grounded in the agent's desires and beliefs.
\end{myDef}
\noindent Motivated by the broad theory of intent~\citep{cushman2015deconstructing}, we decompose intent into three key elements, comprising \textit{beliefs}, \textit{desires}, and \textit{plans}.

\vspace{2mm}
\noindent \textbf{Belief.}
In modern cognitive philosophy, ``belief'' describes an agent's acceptance of a proposition as true or factual~\citep{sep-belief}. This concept does not require active contemplation or accuracy~\citep{armstrong1973belief, wellman2001meta}. For instance, beliefs about vaccination can include propositions such as ``vaccines are detrimental to life,'' ``vaccines work effectively, though they can have side effects,'' or ``vaccination—the most effective method to mitigate the spread of disease.'' Belief evokes a sense of constraint~\citep{pettit1996freedom} while also eliciting interpersonal conflict and discord~\citep{boichak2020digital, sassenberg2024intraindividual}.
\begin{myDef}[Belief]
\label{def:belief}
Belief is a mental condition where a person adopts a particular stance, attitude, or viewpoint regarding a proposition.
\end{myDef}

\noindent \textbf{Desire.}
Contemporary philosophy suggests that moral quality attribution is the inference about an individual's desires~\citep{ames2015perceived}. Desires, as representational cognitive states, convey a positive attitude toward the scenarios they depict~\citep{sinhababu2017humean} and drive related actions~\citep{smith1994moral, anscombe2000intention}. For example, news articles about vaccination may be driven by desires such as ``expressing concerns about vaccine dangers'', ``criticizing government policies'', ``capturing interest for financial gain'', or ``educating the public about vaccination.''
\begin{myDef}[Desire]
\label{def:desire}
To desire ${p}$ means having a tendency to engage in actions thought to facilitate achievement of ${p}$.
\end{myDef}

\noindent \textbf{Plan.}
An individual may have many desires, but without commitment, these desires often remain unacted upon \citep{bratman2009intention}. Commitment requires formulating plans, which involve considering conditions \citep{bratman1987intention} and simulating behavior sequences \citep{phillips2002infants}. By devising specific plans, individuals refine options to identify the components of their final intent \citep{sukthankar2014plan}. For instance, in reporting on vaccination risks, plans might include ``emphasizing risks while presenting both pros and cons,'' ``highlighting severe side effects with doctor quotes,'' or ``fabricating alarming death rates and suggesting a government conspiracy.'' These illustrate deliberate narrative crafting and allow flexibility in responding to changing circumstances~\citep{sheng2022zoom}.
\begin{myDef}[Plan]
\label{def:plan}
Plans are devised to address desires by setting objectives and linking actions to their anticipated outcomes.
\end{myDef}

\subsection{\FrameName Framework} \label{ssec:NINT}

By deconstructing intent into three fundamental components, we aim to further instantiate their presentation in news articles, as news content is the principal conveyor of information to the audience \citep{newcomb1958psychology, van2013news}.
Specifically, we introduce the conceptual deconstruction-based news intent understanding framework, \FrameName, for news intent identification and develop a multi-level and fine-grained classification system, considering the aspects of news writing and social reactions.

\vspace{0.1cm}
\noindent \textbf{Belief identification.}
Beliefs shape the perspective and attitude of news articles, influencing their narrative direction. According to \citet{kuccuk2020stance}, beliefs can be categorized into three classes: \textit{(i) Favor}, \textit{(ii) Against}, and \textit{(iii) Neutral}.

Special attention should be given to non-neutral news articles. While complete neutrality is unrealistic, systematic bias can lead to harmful outcomes like polarization \citep{spinde2021automated}. Identifying beliefs is treated as a multi-class classification task.

\vspace{0.1cm}
\noindent \textbf{Desire identification.} Desires are reflected in the intended impact of the news and the internal as well as external benefits pursued, representing their core purpose. We examine desires \citep{murayama2022annotation} across two facets:
\begin{itemize}[itemsep=2pt,topsep=2pt,parsep=0pt]
    \item[i)] Impact on the target. Intent to induce specific effects on a target, such as harming reputation, giving commendation, or inciting violence \citep{pramanick2021momenta, lu2023facilitating}.
    \item[ii)] Desired social reactions. To provoke public responses, including fostering dialogues \citep{mendoza2020gene}, triggering collective emotions \citep{luvembe2023dual}, and influencing public opinion \citep{mahony2024concerns}.
\end{itemize}

Desires are categorized as follows: \textit{(i) Public interest,} aimed at benefiting the general populace \citep{porlezza2024datafication} (e.g., disseminating scientific information). \textit{(ii) Political interest,} intended to promote a political agenda \citep{son2024teen} (e.g., undermining election contenders). \textit{(iii) Economic interest,} focused on financial gain \citep{immorlica2024clickbait, quandt2024euphoria} (e.g., producing sensational stories for profit). \textit{(iv) Psychological fulfillment,} aimed at seeking pleasure or social interaction \citep{wardle2017information} (e.g., enthusiasm for an ideology, emotional expression, or parody). As news articles can embody multiple desires, the identification of desires is framed as a multi-label classification task.

\vspace{0.1cm}
\noindent \textbf{Plan identification.} Plans are reflected in the writing strategies of news articles, designed to achieve desires based on contextual beliefs \citep{mele1997philosophy, heider1958psychology}. Following \citet{piskorski2023semeval}, we examine two aspects of plans:
\begin{itemize}[itemsep=2pt,topsep=2pt,parsep=0pt]
    \item[i)] News framing. Defined as ``highlighting the prominence of some facets of a topic'' \citep{ali2022survey, sinelnik2024narratives}, we use a classification system with 14 general categories \citep{piskorski2023semeval}.
    \item[ii)] Persuasion. This involves particular linguistic choices aimed at swaying the audience \citep{hasanain2024can, sajwani2024frappe}.
\end{itemize}
The overarching categorization for plan is based on journalism's ethical standards \citep{ireton2018journalism}: \textit{(i) Fair,} news should present diverse viewpoints, aim for equilibrium, and offer adequate context. Fair reporting builds trust and credibility; \textit{(ii) Unfair,} lacking balanced perspectives or sufficient context. Plan identification is framed as a binary classification task.

\vspace{0.1cm}
\noindent \textbf{News intent identification task.} Building on the previous bottom-up analysis, the intent identification task is divided into belief, desire, and plan identification. To assess the societal impact of news and provide a holistic evaluation, intent polarity is categorized as: \textit{(i) Harmful,} which threatens democracy and scientific progress, or exacerbates societal divisions, fosters an environment of increased polarization and hostility among different groups \citep{sheng2022characterizing}; \textit{(ii) Unharmful,} which aims to inform, educate, or entertain without causing harm, foster understanding, and encourage positive discourse.

\section{Dataset} \label{sec:dataset}

\subsection{Data Collection} \label{ssec:collection}

\begin{figure}[htb]
    \centering
    \includegraphics[width=.8\linewidth]{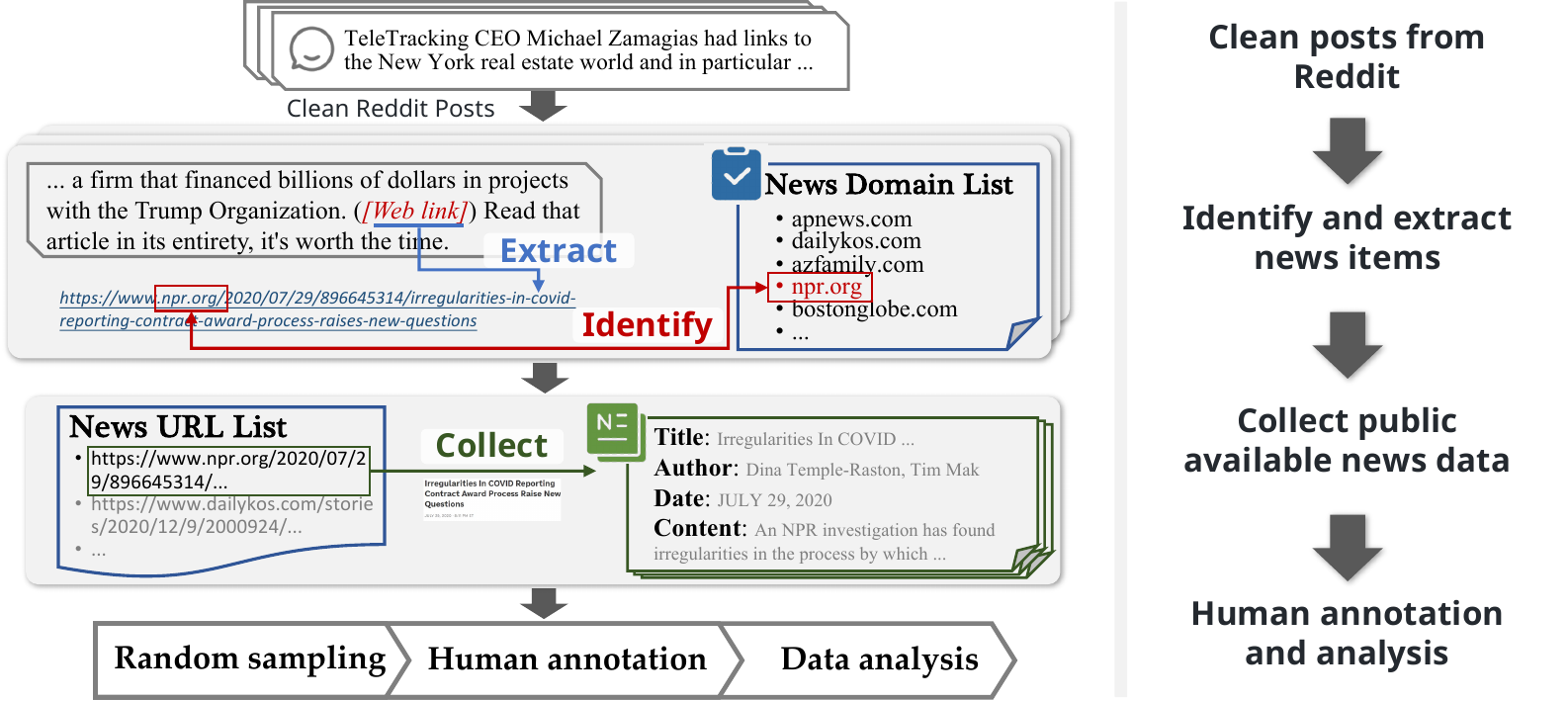}
    \caption{Procedure of \DatasetName dataset curation.}
    \label{fig:process}
\end{figure}

To ensure a semantically rich and diverse dataset with substantial social reactions, we focus on news topics that have generated significant online discussion. We begin with the Factoid \citep{sakketou2022factoid} dataset, which contains 3.4 million Reddit posts discussing contemporary news. As shown in \figureautorefname~\ref{fig:process}, the process starts by cleaning the post content to remove irrelevant elements. We then use regular expressions to identify news links\footnote{\url{https://mediabiasfactcheck.com/}}. These links are manually verified to ensure dataset integrity. Finally, we collect publicly accessible news data from the verified links.

\begin{figure}[ht]
\centering
    \begin{minipage}[b]{0.25\linewidth}
    \centering
    \includegraphics[width=\linewidth]{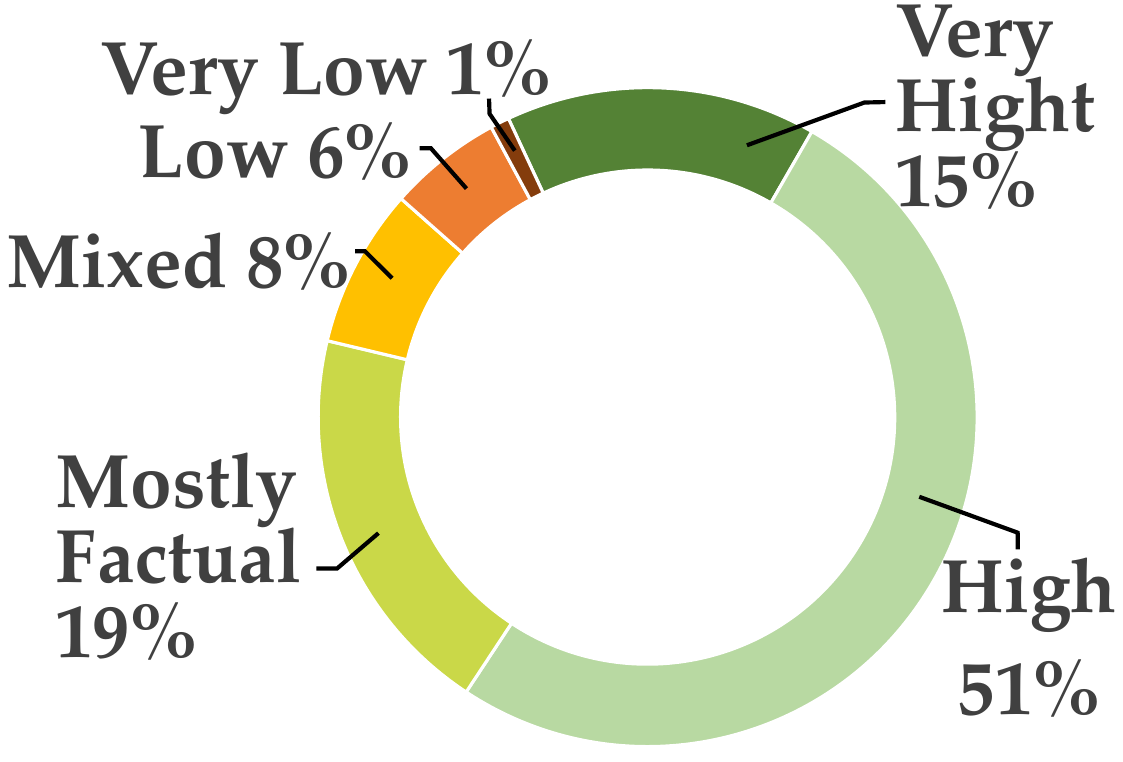}
    \end{minipage}
    \hspace{55pt}
    \begin{minipage}[b]{0.25\linewidth}
    \centering
    \includegraphics[width=\linewidth]{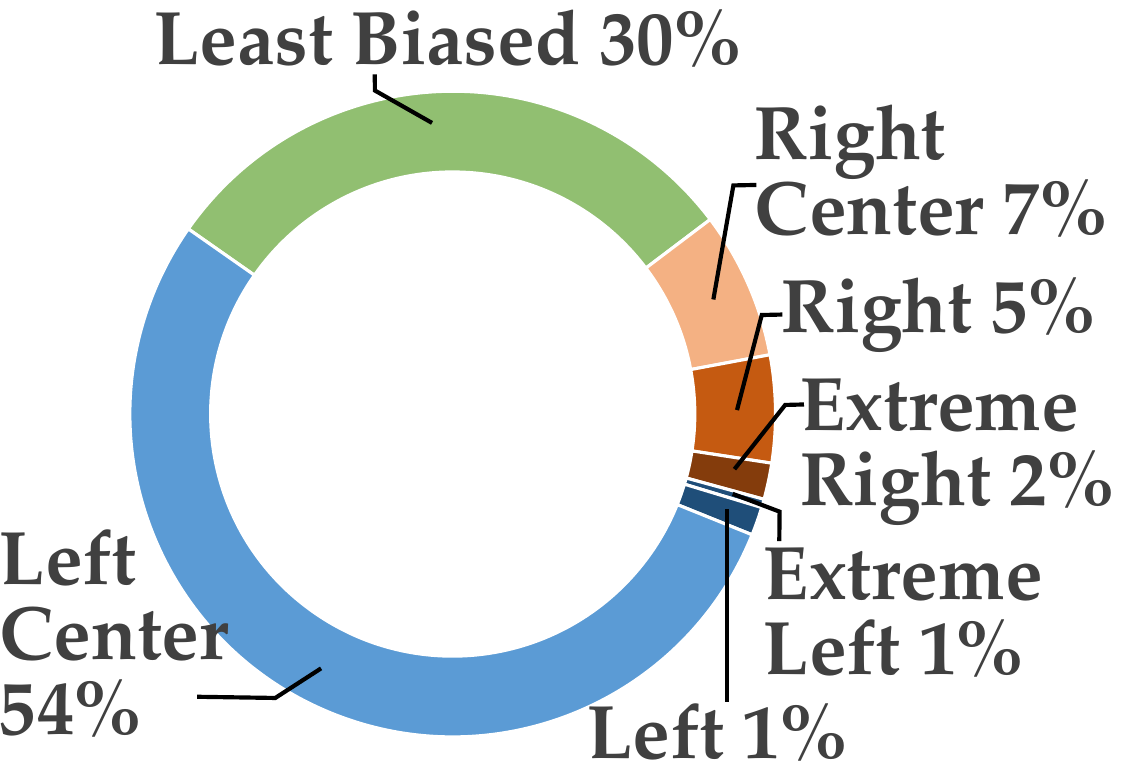}
    \end{minipage}
    \caption{Data distribution in factuality (left) and political bias level (right).}
    \label{fig:distribution}
\end{figure}

In general, \DatasetName comprises 12,959 news articles, featuring a diverse range of news types and a rich social context, with an average of 5.45 related discussion entries per article (detailed statistical information is presented in \tableautorefname~\ref{tab:topic_statistic}). \figureautorefname~\ref{fig:distribution} displays data statistics categorized by factuality and political bias levels.

\begin{table}[H]
    \small
    \renewcommand\arraystretch{1.0}
    \aboverulesep=0.5ex 
    \belowrulesep=0.5ex 
    \centering
    \caption{Statistical information of \DatasetName based on news discussed in different subreddits.}
    \label{tab:topic_statistic}
    \resizebox{.5\linewidth}{!}{
        \begin{tabular}{lrrr}
        \toprule
        \textbf{\centering{Subreddit}}    & \textbf{\centering{\#News}} & \textbf{\centering{Avg. len.}} & \textbf{\centering{Avg. \#posts}} \\ \midrule
        r/Conservative & 676 & 854.51 & 5.63 \\
        r/conservatives & 169 & 889.22 & 6.04 \\
        r/Republican & 461 & 911.07 & 2.23 \\
        r/ConservativesOnly & 89 & 912.63 & 1.34 \\
        r/democrats & 235 & 861.45 & 2.23 \\
        r/uspolitics & 254 & 1063.21 & 1.90 \\
        r/RepublicanValues & 11 & 630.64 & 1.82 \\
        r/Liberal & 157 & 1162.64 & 2.11 \\
        r/JoeBiden & 149 & 936.56 & 6.46 \\
        r/ImpeachTrump & 71 & 751.25 & 1.54 \\
        r/politics & 6884 & 865.30 & 9.08 \\
        r/NoNewNormal & 256 & 862.36 & 6.18 \\
        r/LockdownSkepticism & 243 & 976.01 & 6.63 \\
        r/NoLockdownsNoMasks & 107 & 874.70 & 2.20 \\
        r/Coronavirus & 164 & 788.77 & 4.66 \\
        r/CoronavirusUS & 103 & 1015.86 & 1.45 \\
        r/COVID19 & 47 & 1059.77 & 1.49 \\
        r/EndTheLockdowns & 41 & 837.93 & 1.29 \\
        r/CoronavirusRecession & 27 & 996.85 & 2.22 \\
        r/AntiVaxxers & 27 & 1165.52 & 2.59 \\
        r/Masks4All & 26 & 1086.88 & 3.19 \\
        r/vaxxhappened & 24 & 874.88 & 1.63 \\
        r/COVID19positive & 15 & 1203.40 & 1.53 \\
        r/CoronavirusCanada & 28 & 1120.89 & 2.07 \\
        r/CoronavirusUK & 133 & 597.16 & 2.97 \\
        r/LockdownCriticalLeft & 129 & 962.64 & 2.94 \\
        r/climatechange & 390 & 1290.24 & 4.28 \\
        r/climateskeptics & 267 & 1325.09 & 5.93 \\
        r/MensRights & 254 & 1042.81 & 7.10 \\
        r/Egalitarianism & 30 & 1640.57 & 6.97 \\
        r/antifeminists & 41 & 1053.49 & 1.80 \\
        r/feminisms & 17 & 1076.36 & 1.10 \\
        r/DebateVaccines & 403 & 1097.18 & 9.55 \\
        r/DebateVaccine & 33 & 1357.00 & 2.12 \\
        r/TrueAntiVaccination & 34 & 1182.82 & 3.47 \\
        r/prolife & 137 & 937.11 & 2.86 \\
        r/Abortiondebate & 76 & 970.75 & 6.91 \\
        r/prochoice & 63 & 1074.00 & 2.17 \\
        r/progun & 192 & 862.29 & 2.64 \\
        r/liberalgunowners & 155 & 977.68 & 2.12 \\
        r/Firearms & 143 & 969.31 & 2.01 \\
        r/GunsAreCool & 76 & 856.93 & 1.78 \\
        r/gunpolitics & 51 & 820.27 & 2.02 \\
        r/guncontrol & 50 & 2227.56 & 7.44 \\ 
        r/5GDebate	&21	&1196.10	&2.10 \\
        \bottomrule
        \end{tabular}
    }
\end{table}

\subsection{Data Annotation} \label{ssec:annotation}

We transform our framework into a sequential question-answering process for annotation, following a fine-to-coarse approach. Annotators start with writing strategy analysis (e.g., persuasive language, frame analysis) and progress to high-level intent analysis (e.g., desire identification), concluding with intent polarity classification. Three team members with expertise in mass communication and annotation developed a guideline that includes instructions and examples to clarify each element and aims to provide annotators with clear, well-defined annotation guidelines that minimize the potential for unrestricted subjective interpretations. This guideline balances concise intuition-based and detailed instructions, aiding in recruiting more annotators. Feedback from trial annotations is used to refine the guidelines iteratively. We use Qualtrics\footnote{\url{https://www.qualtrics.com/}} for designing and managing the annotation interface. 

The study has received approval from the research ethics committee, and all data collection has consented to research use. We ensured that the recruited workers were paid fairly and conducted an optional post-study demographics survey. 

\subsubsection{Annotation Procedure}

\begin{figure}[ht]
    \centering
    \includegraphics[width=.9\linewidth]{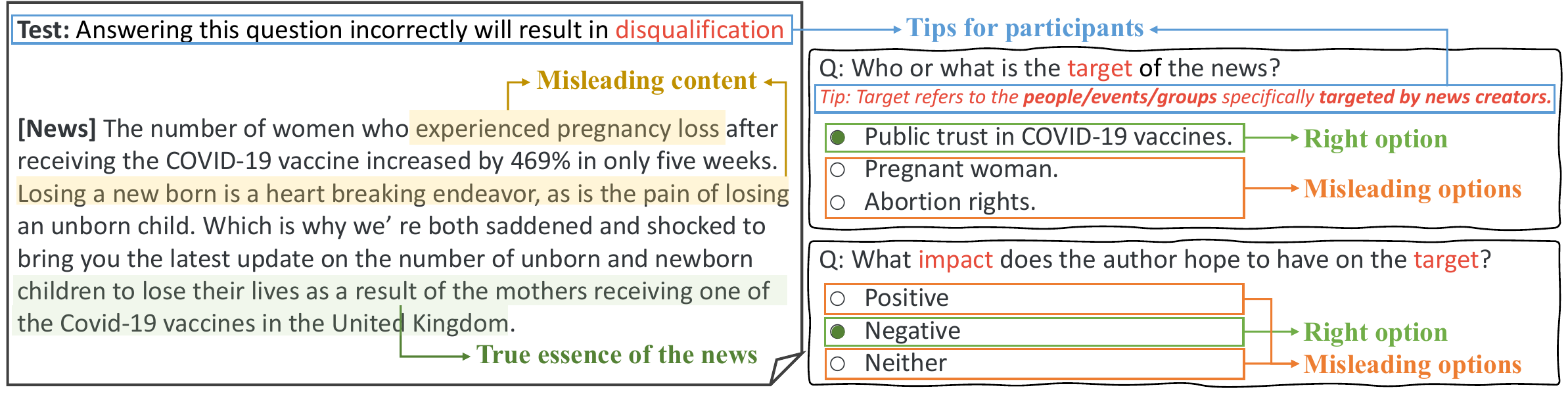}
    \caption{Test questions for \textit{qualification} stage.\label{fig:tutorial}}
\end{figure}

To gain human readers' perception of news intent, considering the news collected is mainly focused on American social life, we recruit native English speakers residing in the USA for data annotation. Simultaneously, we strive to recruit annotators from diverse geographical regions, ethnicities, and socio-economic backgrounds. The recruitment is done via Prolific\footnote{\url{https://app.prolific.com/}}, a platform for high-quality data collection. 

The annotation procedure consists of a \textit{{qualification}} stage and a \textit{{task}} stage. 
During the \textit{{qualification}} stage, participants complete a tutorial to familiarize themselves with the task and interface. This 10-minute, example-centric training session includes: (1) definitions with concise examples, (2) a hands-on approach where annotators achieve a coherent grasp of each concept by providing them with a collection of pre-classified examples of each concept, and (3) two comprehension tests using examples that highlight common challenges in understanding news intent (see \figureautorefname~\ref{fig:tutorial}).

Only annotators who complete the full training and pass both comprehension tests are enlisted into the \textit{{task}} stage, wherein it provides annotations for 20 randomly assigned examples. The recruitment process persists until each news item secures at least three annotations to ensure reliability. Low-quality questionnaires are discarded if either of the two attention checks presented every 10 items fails.

\subsubsection{Quality Control}

To enhance annotation quality, we incorporated feedback from the experimental annotation process. By prioritizing questions regarding the fundamental elements of news writing strategies, annotators are encouraged to read the news more carefully while arranging questions concerning higher-level elements (such as desires and intent polarity) to later sections, following a fine-to-grain sequential annotation strategy. 
 
After conducting trial annotations and discussions, we also found that continuous exposure to the same themes can increase cognitive load and lead to over-sensitivity to patterns. To alleviate this, we reorganized the data to avoid consecutive articles on the same topic, aiming to reduce fatigue by interspersing varied topics, keeping annotators more engaged and less weary.

We also implement a dual quality control procedure. In addition to assessing participants' understanding of news intent and conducting screenings, manual reviews are performed to further ensure the reliability of the annotations: whether the time spent and the number of clicks per annotation page are consistently and significantly below the estimated values during the trial annotation.

Overall, $N=57$ participants joined our project, with $13$ disqualified for failing to meet the criteria as mentioned earlier. Ultimately, over 2,200 fine-grained annotations from multi-topic news articles were included.

\subsubsection{Annotation Results} \label{appen:annotation-results}

\begin{table}[H]
\centering
\renewcommand\arraystretch{1.05}
  \caption{An Instance from the \DatasetName dataset.} \label{tab:data_case}
  \small
  \setlength{\tabcolsep}{5pt}{
  \begin{tabular}{m{12cm}}
    \toprule
    \rowcolor{verylightgray} 
    \textbf{News article} \\
    {\color{bettergreen} \footnotesize[\texttt{title}]} 
    {\fontsize{8.5pt}{-1pt}\selectfont Big Tech’s Domination of Business...} \\
    {\color{bettergreen} \footnotesize[\texttt{content}]}
    {\fontsize{8.5pt}{-1pt}\selectfont American tech titans flew high before the coronavirus pandemic, making billions...
    } \\
    {\fontsize{8.5pt}{-1pt}\selectfont
        {\color{bettergreen} \footnotesize[\texttt{domain}]} nytimes.com \quad
        {\color{bettergreen} \footnotesize[\texttt{date}]} 2020-08-19
        }\\
    {\fontsize{8.5pt}{-1pt}\selectfont
        {\color{bettergreen} \footnotesize[\texttt{author}]} Peter Eavis and ... 
        {\color{bettergreen} \footnotesize[\texttt{URL}]} https://www.nytimes.com/...
        }\\
    \hdashline
    \rowcolor{verylightgray} 
    \textbf{Social context} \\
    {\fontsize{8.5pt}{-1pt}\selectfont
        {\color{bettergreen} \footnotesize[\texttt{discussed in}]} r/LockdownCriticalLeft 
        }\\
    {\fontsize{8.5pt}{-1pt}\selectfont
        {\color{bettergreen} \footnotesize[\texttt{discussed by}]} `gloapg3': `The fact that this has to be explained is what's crazy... (2020-11-18 02:25:16)', ..., `gco9rxf': `Look, it's now perfectly clear that... (2021-02-02 02:01:02)'
        }\\
    \hdashline
    \rowcolor{verylightgray}
    \textbf{Intent label} \\
     {\color{bluev} \footnotesize [\texttt{Writing strategy}]} news framing: economic problem, fairness and equality problem
     \\
     {\color{bluev} \footnotesize [\texttt{Writing strategy}]} persuasion: yes 
     \\
     {\color{bluev} \footnotesize [\texttt{Plan label}]} unfair 
     \\
     {\color{yellowv} \footnotesize [\texttt{Belief label}]} big tech’s domination, against \\
     {\color{redv} \footnotesize [\texttt{Targeted individual/group)}]} tech titans, negative \\
     {\color{redv} \footnotesize [\texttt{Desired social reactions}]} social debate: no \\
     {\color{redv} \footnotesize [\texttt{Desired social reactions}]} opinion shift: no \\
     {\color{redv} \footnotesize [\texttt{Desired social reactions}]} social emotion: surprise, distrust \\
     {\color{redv} \footnotesize [\texttt{Desire label}]} public interest, economic interest \\
     {\color{gray} \footnotesize [\texttt{Intent polarity}]} unharmful \\
     \hdashline
     \rowcolor{verylightgray} 
     \textbf{Domain label} \\ 
     {\color{bettergreen} \footnotesize [\texttt{Veracity}]} high reliability \quad {\color{bettergreen} \footnotesize[\texttt{Bias}]} left center \\
    \bottomrule
  \end{tabular} }
\end{table}

\begin{figure}[t]
\centering
    \begin{minipage}[b]{0.35\linewidth}
        \centering
        \includegraphics[width=\linewidth]{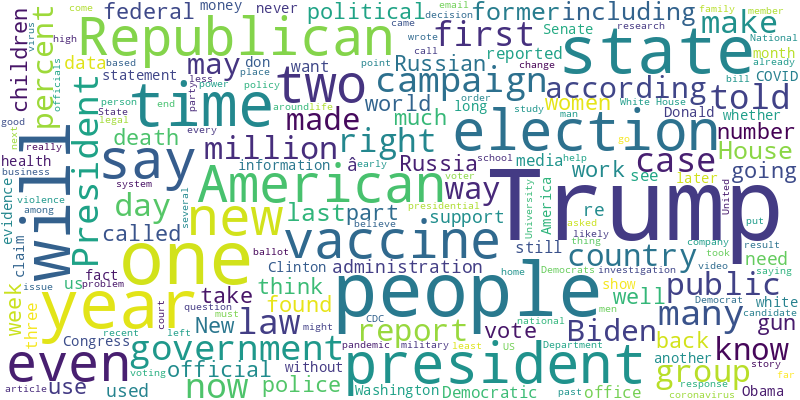}
    \end{minipage}
    \hspace{20pt}
    \begin{minipage}[b]{0.35\linewidth}
        \centering
        \includegraphics[width=\linewidth]{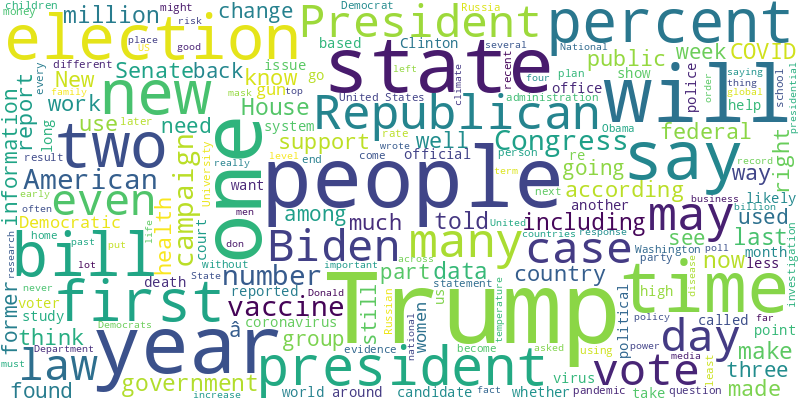}
    \end{minipage}
    \caption{Word clouds of news with harmful intent (left) and unharmful intent (right).}
    \label{fig:word_clouds}
\label{fig:wordcloud}
\end{figure}

An example from the annotated dataset is presented in \tablename~\ref{tab:data_case}. Word clouds of news with harmful (left) and non-harmful (right) intent are illustrated in \figurename~\ref{fig:word_clouds}. While the subjects of harmful and non-harmful news differ slightly, predicting intent based solely on content topics is challenging.

An optional demographics section was included at the end of the survey. Of those who disclosed their gender, 51.3\% identified as male and 48.7\% as female. 
Regarding educational attainment, the majority had completed an undergraduate degree (46.2\%), followed by a graduate degree (33.3\%), technical/community college (7.7\%), high school diploma/A-levels (5.1\%), and doctorate degree (5.1\%). 
In terms of age, most participants were aged 25-50 (76.7\%), with smaller groups aged 18-24 (7.7\%) and 51+ (15.6\%). Employment status varied, with the majority being in full-time work (61.5\%), followed by unemployed but seeking work (15.4\%), part-time work (10.3\%), not in paid work such as homemakers or disabled (7.7\%), and students (5.1\%).

\noindent
\begin{minipage}{0.5\textwidth}
\renewcommand{\thetable}{\arabic{table}}
The inter-annotator agreement scores for intent polarity, belief, desire, and plan annotation using multi-rater kappa \citep{fleiss1971measuring} are presented in \tableautorefname~3, along with the interpretation following \citet{landis1977measurement}. 
Our secondary decomposition of intent compositions and the well-designed annotation guidelines contribute to achieving adequate annotator consistency across these categories.
\end{minipage}
\hfill
\begin{minipage}{0.45\textwidth}
\renewcommand{\thetable}{\arabic{table}}
\noindent
        \centering
        \small
        \captionof{table}{Inter-annotator agreements (Fleiss's $\kappa$).}
        \label{tab:kappa}
        \setlength\tabcolsep{2pt}
        \begin{tabular}{cccc}
        \toprule
         Intent    & Belief    & Desire   & Plan    \\
        \midrule
         \makecell{0.63\\(Substantial)} & \makecell{0.60\\(Moderate)} & \makecell{0.55\\(Moderate)} & \makecell{0.71\\(Substantial)} \\
        \bottomrule
        \end{tabular}
\end{minipage}

\section{Methods}\label{sec:methods}
\subsection{Method for News Intent Identification} \label{ssec:DMG}

\begin{figure*}[htb]
    \centering
    \includegraphics[width=\linewidth]{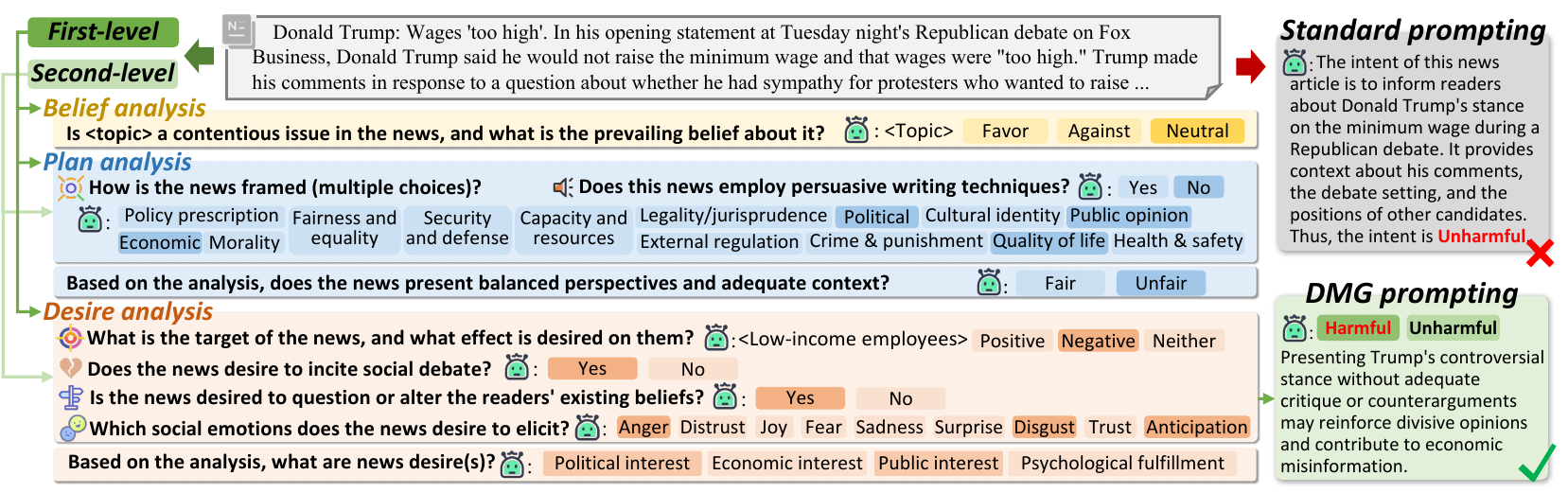}
    \caption{Illustration of \LLMName Prompting.} 
    \label{fig:method-LLM}
\end{figure*}

Computational methods enable scalable news intent analysis for various critical applications and are crucial for the automated comprehension of global news narratives.
However, the development of automated methods relies on costly manual annotation, which creates scalability bottlenecks. To address this issue, we investigate the potential of Large Language Models (LLMs) as cognitive simulators and explore their potential to perceive news intent.

\noindent \textbf{Problem definition.}
Based on the proposed framework \FrameName, we decompose news intent identification into belief, plan, and desire identification sub-tasks, with a special focus on the potential influence of news intent detecting intent polarity at the article level. 

Given an input sequence $x$ representing the content of a news article and a model $\mathcal{M}$, let a task-specific prompt prepended to the input be $p$, and the output is $y$. Under this formulation, Standard Prompting can be described as:
\begin{align}
    \textit{Standard Prompting:} \quad{} & y = \mathcal{M}(x \parallel p),
\end{align}
which constructs prompts that only contain the task description and the given news. 

\noindent \textbf{\LLMName for news intent identification.}
To address the inherent complexity of news intent analysis where surface-level semantics often diverge from latent communicative goals, we propose a \uline{D}econstruction-based \uline{M}ulti-level \uline{G}uidance (\LLMName) Prompting framework. This two-stage cognitive architecture allocates cognitive load, mirroring human workflows while preventing LLM hallucination through constrained reasoning pathways.

\LLMName Prompting is grounded in the \FrameName framework to decompose the intent into constituent elements, breaking down the overall intent into more manageable components (\textit{first-level}), further deconstruction and reasoning of internal intent elements according to its characteristics in the news domain, aiming to guide the model's reasoning from diverse perspectives (\textit{second-level}):
\begin{align}
    \textit{DMG:} \quad{} & \bm{y} = \mathcal{M}(x \parallel \bm{p}_{\text{dmg}}), \\
    & \bm{p}_{\text{dmg}} = \{\bm{p}_\mathcal{B}, \; \bm{p}_\mathcal{P}, \; \bm{p}_\mathcal{D}, \; p\},
\end{align}
where \(\{\mathcal{B}, \mathcal{P}, \mathcal{D}\}\) denote the intent components: belief, plan, and desire, respectively. \(p_{\text{dmg}}\) includes instructions guiding the model \(\mathcal{M}\) through the cognitive reasoning process based on the \FrameName framework. \figureautorefname~\ref{fig:method-LLM} shows a working example. 
 
In the \textit{first-level} deconstruction, intent is deconstructed into belief, plan, and desire, guiding the model to progressively perceive news intent. 
The first component of intent is belief analysis, where the model evaluates the stance of the news regarding the central theme or contentious issue of the news articles, categorizing it as {favor}, {against}, or {neutral}. The second component of intent is plan analysis, the model determines if the news is written fairly with competing perspectives and sufficient context, categorizing it as {fair} or {unfair}. 
To achieve this goal, the \textit{second-level} deconstruction is conducted to guide the model to examine how the news is framed, allowing for multiple selections from categories such as {``Fairness and Equality''} or {``Security and Defense''}, and then assesses whether the article employs persuasive writing techniques. Thirdly, in desire analysis, the goal is broken down by \textit{second-level} deconstruction to identify the intended target of the news and the effect the news aims to achieve, categorizing the effect as {positive}, {negative}, or {neither}. Then, it evaluates whether the news desires to incite social debate, challenge or transform readers' pre-existing beliefs, and identifies the social emotions the news aims to trigger (e.g., {anger}, {trust}, {surprise} etc.). Based on these analyses, the model determines the news's desire, categorizing it into {public interest}, {political interest}, {economic interest}, or {psychological fulfillment}. 

Finally, in the intent polarity analysis, the model synthesizes all the information from the previous analyses to evaluate the potential influence of intent, provides an interpretable polarity scoring disentangled from lexical biases, categorizing it as {harmful} or {unharmful}.

\subsection{Method for News Intent Application}

\begin{figure*}[ht]
    \centering
    \includegraphics[width=.9\linewidth]{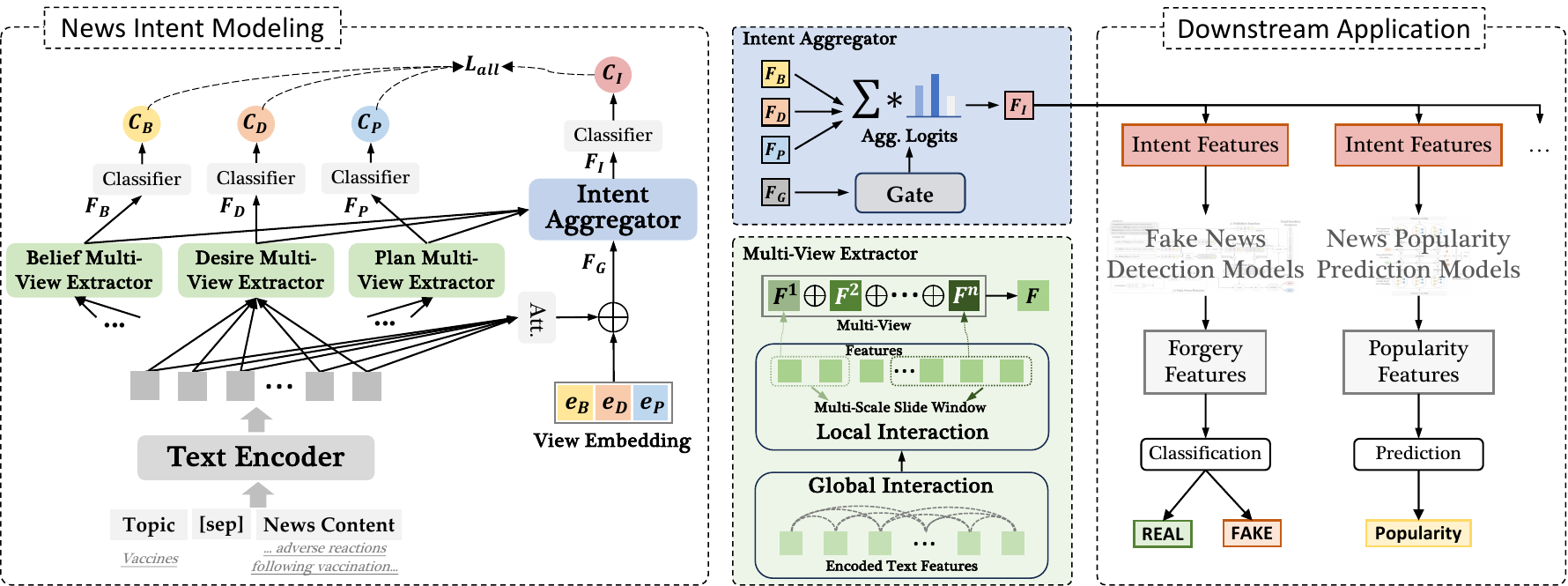}
    \caption{Overview of \ModelName method.}
    \label{fig:application-method}
\end{figure*}

Building on the methods presented in Section~\ref{ssec:DMG}, this section further explores the application of the \FrameName framework in various downstream tasks.
Specifically, we introduce the \uline{D}econstruction-based \uline{M}ulti-view News \uline{INT}ent Modeling (\ModelName) method—a flexible method designed to enhance various models across application tasks. This method benefits from the training signals generated by DMG Prompting while improving efficiency for downstream applications.

As outlined in \figurename~\ref{fig:application-method}, \ModelName introduces a cognitive-inspired architecture that decomposes news intent through multi-view extractors and dynamic view-gated aggregation, enabling joint modeling of compositional semantics and contextual writing patterns via local-global interactions.

\noindent \textbf{Text encoder.}
Given the news topic and news content, we concatenate them using the \([sep]\) token as input. Specifically, RoBERTa \citep{liu2019roberta} is employed as text encoder:
\begin{equation}
\begin{aligned}
    \bm{W} &= \mathrm{RoBERTa}([topic || [sep] || content]),
\end{aligned}
\end{equation}
where $||$ is the concatenation operator.

\noindent \textbf{Multi-View Extractor (MVE).} 
Following \FrameName framework, we develop three multi-view extractors to explicitly capture different dimensions of news intent. Each extractor incorporates both global and local interactions. Specifically, the global-view interaction module examines interactions across the entire context to model the global information necessary for each intent dimension, including overall discourse and context,
\begin{align}
    \hat{\bm{F}} &= \mathrm{\text{Extractor}_{global}}(\bm{W} ;\theta),
\end{align}
where $\theta$ denotes the relevant parameters. Additionally, modeling the content of specific segments and even words is crucial for understanding detailed framing strategies, event progression, and specific writing techniques in the news. A local-view interaction module is introduced, incorporating multiple sliding windows of varying sizes to facilitate multi-scale local interactions,
\begin{align}
    \bm{F^i} &= \mathrm{\text{Extractor}_{local}}(\hat{\bm{F}} ;k_{i}), i \in \{1, 2, ..., n\}, \\
    \bm{F} &= [F^1 || F^2 || ... || F^n], 
\end{align}
where $k_{i}$ represents the parameters within the \(i\)-th sliding window and \(\bm{F^i}\) denotes the corresponding output features. 
By integrating global and local interactions, MVE enhances the comprehensiveness and accuracy of understanding news intent. 
In practice, the global and local interaction modules are implemented using multi-head attention and 1D convolutions with varying kernel sizes.

Given the extracted specific features $\bm F_t$ for each $t \in \{\mathcal{B}, \mathcal{D}, \mathcal{P}\}$, representing belief, desire, and plan, respectively, we can obtain the corresponding identification result $\bm C_t$, as follows
\begin{align}
    \bm{C_t} &= \sigma (\mathrm{MLP}(\bm{F_t})),
\end{align}
where $\sigma(\cdot)$ represents the sigmoid function, $\mathrm{MLP}$ refers to the multi-layer perceptron employed for classification.

\noindent \textbf{Intent Aggregator (IA).}
An intent aggregator is designed to create the ultimate representation $\bm{F_I}$ of news intent. Avoiding neglect of the emphasis on dimension-specific information, the intent aggregator builds an adaptive approach, allowing the proposed \ModelName to adjust and combine these representations dynamically. Given $\bm{W}$ and $\bm{E}$, the weight vector $F_G$ is calculated by
\begin{equation}
\begin{aligned}
    F_G = \mathrm{MAttn}(\bm{W}) \oplus \bm{E},
\end{aligned}
\end{equation}
where $\mathrm{MAttn}(\cdot)$ is a masked attention module for encoding the context features of input news and $\oplus$ denotes the sum operation. $F_G$ is then fed into a gate network to derive the aggregating logits $\bm{M}$ of each dimension:
\begin{equation}
\bm{M} = \mathrm{softmax}(\mathrm{Gate}(F_G; \phi)),
\end{equation}
where $\phi$ stands for the parameter of the gate network, and the softmax function is used to normalize the aggregating logits $\bm{M}$. Finally, the ultimate representation of news intent can be driven by weighting the dimension-specific representations:
\begin{equation}
\begin{aligned}
    \bm{F_I} &= \sum_{t\in \{\mathcal{B}, \mathcal{D}, \mathcal{P}\}} \bm{M_t}\bm{F_t}, \\
    \bm{C_I} &= \sigma (\mathrm{MLP}(\bm{F_I})).
\end{aligned}
\end{equation}

\noindent \textbf{Loss functions.}
The Cross-Entropy Loss ($L$) is employed for the classification:
\begin{equation}
\begin{aligned}
    L(\hat{y}, y) = \!-\!\sum_{k=1}^N(y^k\log \hat{y}^k\!+\!(1\!-\!y^k)\log(1\!-\!\hat{y}^k)),
\end{aligned}
\end{equation}
where $y$, $\hat{y}$, and $k$ represent the ground truth, prediction, and the number of categories, respectively. Given ground truth $\bm{Y}$, $L_{all}$ can be formulated as: 
\begin{equation}
\begin{aligned}
    L_{all} &= \sum_{t\in \{\mathcal{B}, \mathcal{D}, \mathcal{P}, \mathcal{I}\}} L(\bm{C_t}, \bm{Y_t}).
\end{aligned}
\end{equation}

\section{Results} \label{sec:results}

In this section, we present the results of experiments on the intent identification task to demonstrate the performance of our proposed method for understanding news intent. Following this, we conduct experiments on three application tasks, showing that the news intent learned through our approach can enhance model performance in these representative news-related tasks.

\subsection{Results on News Intent Identification}

\begin{table}[htbp] 
\small
\renewcommand\arraystretch{1.3}
\setlength\tabcolsep{10pt}
\caption{\LLMName in news intent polarity classification.} 
\label{tab:DMG_results}
  \centering
    \begin{tabular}{l cccc}
    \toprule
    Method & macF1$\uparrow$ & Prec$\uparrow$ & Rec$\uparrow$ & Acc$\uparrow$ \\
    \midrule
    GPT3
            &  0.438  &  0.459  &  0.449  &  0.475
            \\
    \ \ \ \ w/ \LLMName
            &  \textbf{0.484}  &  \textbf{0.532}  &  \textbf{0.538}  &  \textbf{0.492}
            \\
    T5-XXL           
            &  0.451  &  0.449  &  0.454  & 0.575
            \\
    \ \ \ \ w/ \LLMName
            &  \textbf{0.552}  &  \textbf{0.553}  &  \textbf{0.552}  & \textbf{0.645}
            \\
    QwQ-32B
            & 0.518 & 0.656 & 0.544 & 0.720 
            \\
    \ \ \ \ w/ \LLMName
           & \textbf{0.604} & \textbf{0.667} & \textbf{0.598} & \textbf{0.741}
            \\
    LLaMA-13B
            &  0.585  &  0.628  &  0.582  &  0.717
            \\
    \ \ \ \ w/ \LLMName
            &  \textbf{0.611}  &  \textbf{0.687}  &  \textbf{0.602}  & \textbf{0.753}
            \\
    LLaMA-70B         
            &  0.603  & 0.613  & 0.639  &  0.630
            \\
    \ \ \ \ w/ \LLMName
            &  \textbf{0.674}  &  \textbf{0.680}  &  \textbf{0.723}  &  \textbf{0.695}
               \\       
    GPT3.5
            &  0.674  &  0.651  &  0.701  &  0.693
            \\
    \ \ \ \ w/ \LLMName
            & \textbf{0.841} & \textbf{0.824} & \textbf{0.875} & \textbf{0.860}
            \\         
    \bottomrule
    \end{tabular}
\end{table}%

We use intent polarity classification to measure the model's perception of intent, as it requires a comprehensive analysis of news intent. Representative LLM models are analyzed: 
\begin{itemize}[itemsep=2pt,topsep=1pt,parsep=0pt,leftmargin=10pt]
    \item {T5-XXL} refers to \texttt{Flan-T5-XXL} \citep{chung2024scaling}, a large T5 model optimized through instruction tuning for enhanced performance across diverse NLP tasks.
    \item LLaMA-13B and LLaMA-70B refer to \texttt{LLaMA-2-Chat-13B} as well as \texttt{LLaMA-2-Chat-70B} \citep{touvron2023llama}, respectively. These are chat-focused LLaMA variants featuring \{13, 70\} billion parameters for language modeling.
    \item GPT-3 and GPT-3.5 refer to \texttt{GPT3-Davinci} \citep{brown2020language} and \texttt{GPT3.5-turbo} \citep{chatgpt}, respectively. They are representative generative models known for their exceptional language understanding and response generation capabilities.
    \item QwQ-32B refers to \texttt{QwQ-32B-250305} \citep{qwq32b}, a reasoning model based on the Qwen2.5-32B model \citep{qwen2.5}. Compared to conventional instruction-tuned models, QwQ, which possesses advanced thinking and reasoning capabilities, has demonstrated significantly improved performance in many downstream tasks.
\end{itemize}

\subsubsection{Experimental Performance}
The experimental results presented in \tablename~\ref{tab:DMG_results} indicate that the performance of all models has significantly improved through the application of our proposed method\footnote{All metrics are calculated using the Python package scikit-learn.}. With the power of \FrameName, models are endowed with cognitive-level modeling and reasoning abilities. This approach fundamentally reduces reliance on costly human-annotated datasets, enhancing intent identification through structured reasoning patterns that mimic human perception, enabling efficient and human-granular analysis of news ecosystems at scale.

This approach enables a structured and accurate perception of news intent. As shown in \figureautorefname~\ref{fig:method-LLM}, standard prompting falls short due to superficial understanding, while the proposed method provides multi-level, multifaceted information and a deeper perception of news intent.
It is noteworthy that \LLMName prompting requires only a single inference, as questions are sequentially listed, and the model responds to them in order. In the illustration, pairing the answers with their respective questions is intended to enhance readability.

\subsubsection{Cost-Benefit Analysis}

\begin{table}[htbp] \small
\renewcommand\arraystretch{1.2}
\setlength\tabcolsep{2pt}
    \caption{Comparative analysis. \LLMName has significant performance gains and interpretability with moderate computational overhead.}
    \label{tab:llm_tradeoff}
    \centering
    \begin{tabular}{lccccccc}
    \toprule
    \textbf{Method} & {\textbf{\#Queries Needed}} & {\textbf{Avg.Tokens}} & {\textbf{Interpretability}} & {\textbf{macF1}}\\
    \midrule
    
    Standard Prompting & 1 & 104 & \textbf{\redxmark} & 0.674 \\
    Direct CoT & 1 & 489 & {Partial} & 0.710 \\
    Standard CoT & 2 & 988 & {Partial} & 0.722 \\
    \rowcolor{verylightgray}
    \LLMName (Ours) & 
    1 & 213 & \color{teal}\textbf{\checkmark} & 0.841 \\
    \bottomrule
    \end{tabular}
\end{table}

Our framework demonstrates an optimal balance between computational efficiency and task performance through structured reasoning guidance, as quantitatively validated in \tablename~\ref{tab:llm_tradeoff}, with performance exemplified by our method on GPT-3.5.
While standard Chain of Thought (CoT) consumes an average of 988 tokens (2× queries) to reach 0.722 in macF1, \LLMName achieves a 24.8\% relative improvement in macF1 with only an average of 213 tokens (1× query) – demonstrating that unconstrained reasoning paths in CoT lead to 4.6× token inflation with inferior results. The explicit structural constraints in \LLMName eliminate redundant explorations, allowing focused computation.

The token-to-performance ratio further validates this advantage: \LLMName achieves a macF1 gain of 0.153 per 100 tokens, compared to 0.009 for Direct CoT and 0.005 for Standard CoT, indicating 17× higher reasoning efficiency while providing complete interpretability traces. This improvement arises from the shortcomings of CoT's free-form analysis, which lacks cognitive theoretical guidance and explicit logical dependency tracking to minimize cognitive divergence. 

\subsection{Results on News Intent Application}
\begin{table}[htbp]
\small
\renewcommand\arraystretch{1.0}
  \centering 
  \caption{Dataset statistics of the application tasks.}
  \label{tab:application_dataset}
    \begin{tabular}{lllrrr}
    \toprule
    Dataset & Task & Count & Train & Val   & Test  \\
    \midrule
    \multicolumn{1}{l}{\multirow{2}{*}{\rotatebox{0}{\textit{GossipCop}}}} & \multicolumn{1}{l}{\multirow{2}{*}{\rotatebox{0}{Fake News Detection}}} 
        & \#Fake  & 2,024 & 604   & 601 \\
        &  & \#Real  & 5,039 & 1,774 & 1,758 \\
    \midrule
    \multicolumn{1}{l}{\multirow{1}{*}{\rotatebox{0}{\textit{MIND}}}} & News Popularity Prediction 
        &  \#News  & 40,000 & 5,000   & 5,000 \\
    \midrule
    \multicolumn{1}{l}{\multirow{2}{*}{\rotatebox{0}{\textit{QPROP}}}} & \multicolumn{1}{l}{\multirow{2}{*}{\rotatebox{0}{{Propaganda Detection}}}}
        & {\#Prop.}        & {4,015}  & {573}   & {1,149} \\
        &  & {\#Non-prop.} & {31,889} & {4,555} & {9,113} \\
    \bottomrule
    \end{tabular}
\end{table}%

We explore the practical value of news intent in application tasks that benefit from a deeper understanding of news—from what is written to why it is written—news intent can assist in revealing reporting motivations, predicting potential impacts, and enhancing resilience to misinformation and persuasive content. Experiments demonstrate that the intent features captured through our method \ModelName possess noticeable potential to improve the performance of various downstream models.

\noindent \textbf{Application tasks.}
We conduct a series of experiments on three application tasks that benefit from understanding news intent and hold unique research value:
\begin{itemize}[itemsep=2pt,topsep=1pt,parsep=0pt,leftmargin=10pt]
    \item {Fake news detection:} A classification task to distinguish fake news from real news \citep{sheng2021integrating, zhu2022memory, hu2023learn, hu2024bad, nan2024let}, which is essential for maintaining information integrity \citep{zhou2020survey}. Uncovering the deceptive intents behind the news will deepen detectors' news understanding and eventually benefit the judgments.    
    \item {News popularity prediction:} A regression task that holds intrinsic value as a computational proxy for understanding real-world information propagation dynamics \citep{fan2021news}. It predicts popularity by click counts to reflect audience engagement. 
    \item {Propaganda detection: A classification task aims to distinguish propagandistic news from non-propagandistic news. Propaganda refers to the expression of opinion or action by agents that is deliberately designed to influence the opinions or actions of others for predetermined ends \citep{jowett2018propaganda}.
    }
\end{itemize}

\noindent\textbf{Datasets.} The dataset statistics employed for the {three} application tasks are presented in \tableautorefname~\ref{tab:application_dataset}. The \textit{GossipCop} dataset, part of the FakeNewsNet repository \citep{shu2020fakenewsnet}, is divided into training, validation, and test sets based on a temporal partition following prior work \citep{zhu2022generalizing, nan2025exploiting}, simulating real-world fake news detection scenarios. The \textit{MIND} dataset \citep{wu2020mind} is employed adhering to \citet{fan2021news}. 
The \textit{QPROP} dataset \citep{barron2019proppy} consists of 51.3k articles from 104 news sources, with each article labeled as either propagandistic (Prop.) or non-propagandistic (Non-prop.). A temporal split of the dataset is implemented, ensuring that the publication dates of the news in the test set are later than those in the training set, as the model should not have access to future news during training.

\noindent\textbf{Application task-specific baselines.} To evaluate the performance of our inferred news intent features and the proposed method \ModelName, we select corresponding baseline models for three application tasks. The preprocessing, training, and testing configurations follow the corresponding official implementations. We employ late fusion of the intent features with the task features obtained from the baseline model using a two-layer MLP.
For the fake news detection task, we select three representative and recent fake news detection methods as baselines:
\begin{itemize}[itemsep=2pt,topsep=1pt,parsep=0pt,leftmargin=18pt]
    \item BERT-Emo~\citep{zhang2021mining} integrates the writer's emotion into consideration. To ensure fairness in comparisons with other models, we retain only its content-based branch for fake news detection.
    \item ENDEF~\citep{zhu2022generalizing} proposes an entity debiasing method aimed at mitigating entity bias in fake news detection, following a cause-and-effect causal perspective.
    \item MSynFD \citep{xiao2024msynfd} introduces a multi-hop syntactic dependency graph to model syntax information together with sequence-aware semantic information.
\end{itemize}

For news popularity prediction, to access our model's performance, we select the three top-performing methods identified in~\citet{fan2021news} as baseline models:
\begin{itemize}[itemsep=2pt,topsep=1pt,parsep=0pt,leftmargin=18pt]
    \item SeqSpa extracts both sequential and spatial characteristics of news to capture the nuanced patterns effectively.
    \item GCN-buzz leverages the relational structure of news by utilizing a heterogeneous graph and encodes the information flow from the current buzzword to the context.
    \item F$^4$ captures the local as well as global news-entity correlations and designs an aggregation module to provide a holistic representation of news.
\end{itemize}

For propaganda detection, we compare the improvements achieved by our proposed method against the following baseline models:
\begin{itemize}[itemsep=2pt,topsep=1pt,parsep=0pt,leftmargin=18pt]
    \item IPD, the document-level prediction variant proposed by \citet{yu2021interpretable}, is built upon the features identified by \citet{barron2019proppy}, which encompass lexicon, readability, vocabulary richness, and style, and contributes to an effective propaganda detection model.
    \item RoBERTa \citep{liu2019roberta}. We employ a pre-trained RoBERTa model to encode news articles, fine-tuning only the last layer to leverage its enhanced contextual representation capabilities. The news embeddings are then fed into a two-layer multi-layer perceptron for propaganda detection.
    \item DET \citep{chernyavskiy2024unleashing}, a discourse-enhanced transformer-based architecture for propaganda detection, which utilizes Rhetorical Structure Theory for discourse representations.
\end{itemize}

\subsubsection{Experimental Details}
\noindent \textbf{Data preparation.}
To generate sufficient training data for \ModelName, we employ \LLMName Prompting to scale up annotation efforts on the \DatasetName dataset. The structured reasoning procedure not only bypasses the scalability bottlenecks of manual annotation but also establishes a foundational infrastructure for fully automated intent identification systems, representing a promising direction. 
\begin{table}[H]
    \small
    \renewcommand\arraystretch{1.1}
    \setlength{\tabcolsep}{8pt}
      \centering 
      \caption{Dataset statistics.}
        \begin{tabular}{cl rrr r}
        \toprule
        \multicolumn{2}{c}{Count} & Train & Val & Test & Overall \\
        \midrule
        \multicolumn{1}{c}{\multirow{2}{*}{\rotatebox{0}{Intent}}} 
              & \#harmful & 4,911  & 188  & 56  & 5,155   \\
              & \#unharmful & 7,348  & 312  & 144  & 7,804   \\
        \midrule
        \multicolumn{1}{c}{\multirow{3}{*}{\rotatebox{0}{Belief}}}
              & \#favor & 638  & 22  & 17  & 677   \\
              & \#against & 2,473  & 87  & 57  & 2,617   \\
              & \#neutral & 9,148  & 391  & 126  & 9,665  \\
        \midrule
        \multicolumn{1}{c}{\multirow{4}{*}{\rotatebox{0}{Desire}}}
              & \#pub. int. & 3,892  & 174  & 74  & 4,140  \\
              & \#eco. int. & 6,398  & 249  & 32  & 6,679   \\
              & \#pol. int. & 7,847  & 298  & 112  & 8,257  \\
              & \#psy. ful. & 3,043  & 121  & 22  & 3,186 \\
        \midrule
        \multicolumn{1}{c}{\multirow{2}{*}{\rotatebox{0}{Plan}}}
              & \#unfair & 7,840  & 301  & 143  & 8,198   \\
              & \#fair & 4,419  & 199  &  57  & 4,761   \\
        \bottomrule
        \end{tabular}\label{tab:dataset_statistic}
\end{table}%

The best-performing model from Section~\ref{sec:methods} is used to annotate the remaining data in \DatasetName, yielding multi-dimensional fine-grained intent labels (categories of belief, desire, plan, and intent polarity), along with additional labels from the second deconstruction. This enhances the LLM's intent perception and provides more material for analysis and future research. Utilizing LLMs for labeled data augmentation significantly reduces the cost of acquiring supervised data \citep{zhou2023cobra} and has been shown to provide even better labeling quality than crowdsourced human annotations that lack adequate design and quality control \citep{Gilardi2023ChatGPTOC, sheng2025confusing}. This approach creates opportunities for training specialized models for application tasks \citep{ye2025unide}.

\noindent \textbf{LLM-annotated data verification.}
We further conducted human verification on the annotated data generated by our method to assess its capability for augmenting training data. We randomly selected 500 labeled entries for quality assessment. After completing a pilot annotation experiment, all annotators were asked to evaluate whether the provided labels aligned with their understanding of the news (credible/incredible). Fifteen annotators participated in assessing the credibility of news annotation results, and each article was evaluated by at least three independent annotators. Ultimately, 94\%, 97\%, 95\%, and 91\% of the data were judged credible by at least one annotator for beliefs, plans, desires, and intent polarity, respectively. Furthermore, 76\%, 77\%, 74\%, and 78\% samples corresponding to the belief, plan, desire, and intent polarity labels, were collectively labeled as credible by at least three annotators. The verification results demonstrate substantial agreement, with 84\%, 85\%, 83\%, and 86\% pairwise agreements, as well as multi-rater kappa values \citep{Randolph2005FreeMarginalMK} of 0.764, 0.776, 0.749, and 0.795 for beliefs, plans, desires, and intents, respectively. It signifies that the posterior consistency of model annotations with human perception is quite strong.

\noindent \textbf{Data analysis.}
The resulting data distribution is presented in \tablename~\ref{tab:dataset_statistic}. The human-annotated set is designated as the test set, the human-validated set serves as the validation set, and the remaining data is used for model training. In the intent category, the dataset exhibits a significant imbalance, which is consistent with the results of human annotation and reflects the current digital news information landscape \citep{zhang2020overview}. While most news articles exhibit a neutral belief (``neutral'' being the predominant class) by avoiding explicit leaning, they rarely adopt a balanced multi-perspective approach, resulting in an unbalanced view for readers \citep{asp2007fairness} (``unfair'' is the predominant class of plan). 
Furthermore, ``political interests'' emerge as the prevalent category in desire, surpassing the proportion of political news itself. This underscores the pivotal role of news in contemporary democratic political systems. Additionally, the prominence of ``economic interests'' reflects the evolving profit models within the news industry \citep{nelson2024money}.

\begin{figure}[htb]
\centering
    \includegraphics[width=\linewidth]{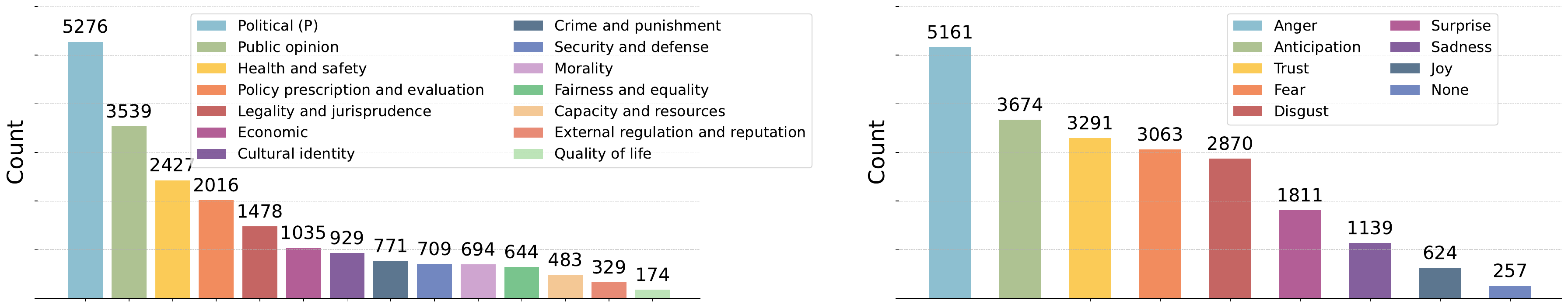}
    \caption{Distribution of news frames and desired social emotion in overall \DatasetName. Left: distribution of various news frames used in news content. Right: distribution of desired social emotion perceived by readers.}
    \label{fig:distributions_analysis}
\end{figure}

Data analyses also reveal a complex interplay between news writing strategy and desired audience engagement. As shown in \figurename~\ref{fig:distributions_analysis}, a strategic approach of media outlets is observed to engage audiences through emotionally charged content, underscoring the power of media in shaping social narratives and the importance of information literacy among audiences. The media's emphasis on political and public opinion frames reflecting the critical role of news framing in shaping public opinion and influencing civic engagement, coupled with the targeting of strong emotions like anger and anticipation, highlights a calculated approach to influencing public discourse and maintaining audience interest. 

\subsubsection{Experimental Performance}

\begin{table}[htbp] \small
\renewcommand\arraystretch{1.}
\setlength\tabcolsep{7pt}
  \centering
  \caption{Ablation studies of modules and subtasks of \ModelName.\label{tab:ablation}}
    \begin{tabular}{lllll}
    \toprule
    Method & macF1$\uparrow$ & Prec$\uparrow$ & Rec$\uparrow$ & Acc$\uparrow$ \\
    \midrule
    \ModelName & \textbf{0.741}  &  \textbf{0.731}  &  \textbf{0.762}  & \textbf{0.770}\\
    \midrule
    w/o MVE &  0.695  &  0.688  &  0.725  &  0.725 \\
    w/o IA  &  0.713  &  0.705  &  0.741  &  0.745 \\
    \midrule
    w/o $\mathcal{B}$      &  0.727  &  0.724  &  0.761  &  0.750 \\  
    w/o $\mathcal{D}$      &  0.719  &  0.711  &  0.755  &  0.745 \\
    w/o $\mathcal{P}$      &  0.730  &  0.721  &  0.747  &  0.765 \\
    \bottomrule
    \end{tabular}%
\end{table}

For \ModelName, the Adam optimizer \citep{kingma2015adam} is used with a learning rate of \(2e^{-5}\), and the batch size is 64. Two-layer MLPs serve as classification heads, with a hidden dimension of $384$ and a ReLU activation.

\noindent \textbf{Effectiveness of \ModelName’s design.}
Ablation experiments are conducted to validate the design of \ModelName. The results on intent polarity classification averaged over five runs are shown in \tableautorefname~\ref{tab:ablation}. The full \ModelName model achieves the highest performance, indicating that both the multi-view extractor (MVE) and intent aggregator (IA) modules contribute significantly to the model's overall performance. To assess the collaborative effects among subtasks, three model variants are introduced: w/o~\(\mathcal{B}\), \(\mathcal{P}\), and \(\mathcal{D}\), each excluding contribution of corresponding subtasks by omitting their influences on the total loss. Results presented in \tableautorefname~\ref{tab:ablation} highlight the importance of every internal element in constructing a nuanced concept of intent and affirm the effectiveness of our proposed framework. 

\noindent {\textbf{Enhancements in application performance.}
Experiments on three application tasks demonstrate that the proposed method can be conveniently deployed alongside various methods, enhancing their performance in different application scenarios.
The average results over five runs are recorded in \tablename~\ref{tab:applications}. Incorporating intent features through our method yields performance improvements. Notably, in fake news detection \ModelName achieves an averaged 2.8\% relative improvement in F1$_{\text{fake}}$, helping enhance fake news mitigation. Similar gains are observed in news popularity prediction and propaganda detection. These consistent improvements highlight our framework's generalizability, providing an effective mechanism for enhancing existing models.

\begin{table*}[htbp] \scriptsize
\centering
\renewcommand \arraystretch{1.4}
\setlength\tabcolsep{1pt}
  \caption{Experimental results of our proposed method \ModelName across three application tasks. The {$\pm$} values denote the standard error of the mean. The results indicated by * are statistically significant compared to its baseline model (p-value < 0.05).}
  \label{tab:applications}
  \begin{tabular}{
    m{1.5cm}m{1.4cm}m{1.4cm}m{1.4cm}
    m{1.5cm}m{1.4cm}m{1.4cm}m{1.4cm}
    m{1.5cm}m{1.4cm}m{1.4cm}m{1.4cm}}
    \toprule
    \multicolumn{4}{c}{\small {Fake news detection}}
    & \multicolumn{4}{c}{\small {News popularity prediction}}
    & \multicolumn{4}{c}{\small {Propaganda detection}}\\
    
    \cmidrule(rl){1-4} \cmidrule(rl){5-8} \cmidrule(rl){9-12}    
    \footnotesize \centering Method
    & \footnotesize \centering macF1$\uparrow$
    & \footnotesize \centering Acc$\uparrow$
    & \footnotesize \centering F1$_{\text{fake}}\uparrow$
    & \footnotesize \centering Method
    & \footnotesize \centering RMSE$\downarrow$
    & \footnotesize \centering MedAE$\downarrow$
    & \footnotesize \centering MAE$\downarrow$
    & \footnotesize \centering Method
    & \footnotesize \centering macF1$\uparrow$
    & \footnotesize \centering Recall$\uparrow$
    & {\footnotesize \centering Prec$\uparrow$} \\
    \hline
    BERT-Emo
    & .787\textsubscript{\color{grayv}$\pm$.0013} 
    & .845\textsubscript{\color{grayv}$\pm$.0011}
    & .683\textsubscript{\color{grayv}$\pm$.0013} 
    &
    SeqSpa 
    & .034\textsubscript{\color{grayv}$\pm$.0008}
    & .006\textsubscript{\color{grayv}$\pm$.0010}
    & .010\textsubscript{\color{grayv}$\pm$.0014}
    &
    IPD
    & .921\textsubscript{\color{grayv}$\pm$.0003} 
    & .901\textsubscript{\color{grayv}$\pm$.0006}
    & .937\textsubscript{\color{grayv}$\pm$.0006} 
    \\
    
    \; + \ModelName
    & \textbf{.813}\textsubscript{\color{grayv}$\pm$.0003}$^*$ 
    & \textbf{.868}\textsubscript{\color{grayv}$\pm$.0007}$^*$ 
    & \textbf{.714}\textsubscript{\color{grayv}$\pm$.0011}$^*$ 
    &
    \; + \ModelName
    & \textbf{.020}\textsubscript{\color{grayv}$\pm$.0009}$^*$ 
    & \textbf{.003}\textsubscript{\color{grayv}$\pm$.0011}$^*$ 
    & \textbf{.008}\textsubscript{\color{grayv}$\pm$.0013}$^*$ 
    &
    \; + \ModelName
    & \textbf{.931}\textsubscript{\color{grayv}$\pm$.0004}$^*$ 
    & \textbf{.925}\textsubscript{\color{grayv}$\pm$.0003}$^*$ 
    & \textbf{.942}\textsubscript{\color{grayv}$\pm$.0002}$^*$ 
    \\

    ENDEF 
    & .798\textsubscript{\color{grayv} $\pm$.0010} 
    & .853\textsubscript{\color{grayv} $\pm$.0006} 
    & .696\textsubscript{\color{grayv} $\pm$.0006} 
    &
    GCN-buzz  
    & .036\textsubscript{\color{grayv} $\pm$.0002} 
    & .005\textsubscript{\color{grayv} $\pm$.0009}   
    & .015\textsubscript{\color{grayv} $\pm$.0013}   
    &
    RoBERTa
    & .902\textsubscript{\color{grayv} $\pm$.0004} 
    & .883\textsubscript{\color{grayv} $\pm$.0003} 
    & .924\textsubscript{\color{grayv} $\pm$.0003}  
    \\    
    
    \; + \ModelName
    & \textbf{.811}\textsubscript{\color{grayv} $\pm$.0007}$^*$ 
    & \textbf{.862}\textsubscript{\color{grayv} $\pm$.0015}$^*$ 
    & \textbf{.712}\textsubscript{\color{grayv} $\pm$.0006}$^*$ 
    & 
    \; + \ModelName
    & \textbf{.025}\textsubscript{\color{grayv} $\pm$.0001}$^*$
    & \textbf{.002}\textsubscript{\color{grayv} $\pm$.0011}$^*$
    & \textbf{.007}\textsubscript{\color{grayv} $\pm$.0010}$^*$
    &  
    \; + \ModelName
    & \textbf{.928}\textsubscript{\color{grayv} $\pm$.0005}$^*$ 
    & \textbf{.904}\textsubscript{\color{grayv} $\pm$.0003}$^*$ 
    & \textbf{.958}\textsubscript{\color{grayv} $\pm$.0004}$^*$ 
    \\ 
    
    MSynFD  
    & .813\textsubscript{\color{grayv}$\pm$.0004} 
    & .857\textsubscript{\color{grayv}$\pm$.0010}
    & .712\textsubscript{\color{grayv}$\pm$.0009}
    &
    F$^4$ 
    & .027\textsubscript{\color{grayv} $\pm$.0003}
    & .003\textsubscript{\color{grayv} $\pm$.0002}
    & .008\textsubscript{\color{grayv} $\pm$.0001}
    &
    {DET} 
    & .936\textsubscript{\color{grayv} $\pm$.0004} 
    & .925\textsubscript{\color{grayv} $\pm$.0007} 
    & .949\textsubscript{\color{grayv} $\pm$.0005}
    \\
    
    \; + \ModelName
    & \textbf{.826}\textsubscript{\color{grayv} $\pm$.0005}$^*$ 
    & \textbf{.878}\textsubscript{\color{grayv} $\pm$.0009}$^*$ 
    & \textbf{.725}\textsubscript{\color{grayv} $\pm$.0008}$^*$
    & 
    \; + \ModelName
    & \textbf{.018}\textsubscript{\color{grayv} $\pm$.0003}$^*$ 
    & \textbf{.002}\textsubscript{\color{grayv} $\pm$.0001}$^*$
    & \textbf{.006}\textsubscript{\color{grayv} $\pm$.0001}$^*$
    &    
    {\; + \ModelName}
    & \textbf{.944}\textsubscript{\color{grayv} $\pm$.0003}$^*$ 
    & \textbf{.934}\textsubscript{\color{grayv} $\pm$.0005}$^*$ 
    & \textbf{.954}\textsubscript{\color{grayv} $\pm$.0003}$^*$
    \\
    \bottomrule
  \end{tabular}
\end{table*}

\section{Discussion} \label{sec:discussion}

\subsection{Internal and External Connection of News Intent}
We provide a preliminary analysis of the proposed \DatasetName dataset, aiming to enhance the understanding of news intent for future research. We offer an analysis of the relationships among the internal elements of news intent, as well as an examination of the connections between the elements of news intent and real-world impacts triggered by the news article.
We show that (1) the internal elements of intent exhibit synergistic consistency, collectively forming the intent behind news production action, which mirrors the theory of rational action \citep{mele1997philosophy}, and (2) within intentional action, intent, actions, and outcomes constitute a causal chain, a causality that is reflected in the relationship between the intent of news production and the social feedback it generates.

\begin{figure}[ht]
    \centering
    \includegraphics[width=.8\linewidth]{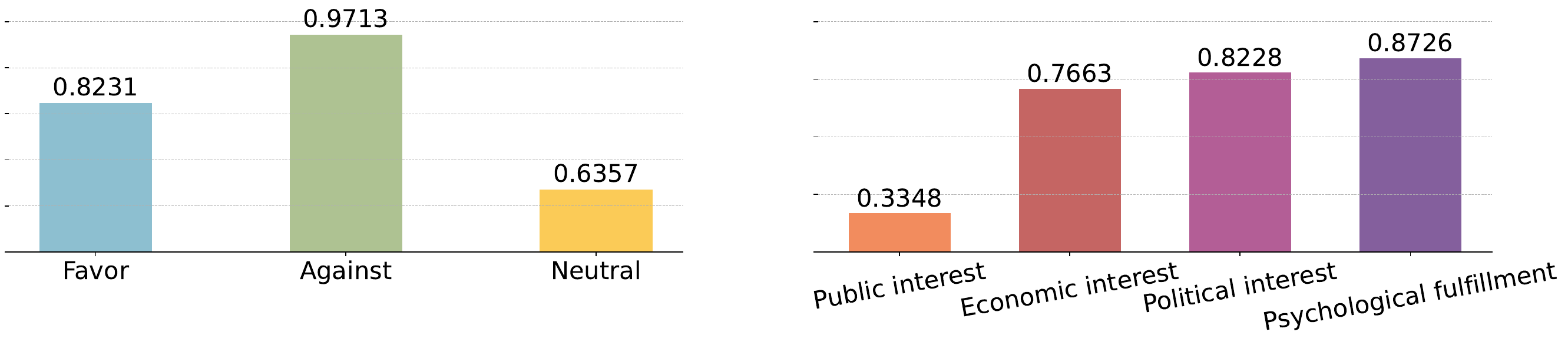}
        \caption{Belief-Plan and Desire-Plan correlation. The bars represent the proportions of news articles with unfair writing plans associated with different beliefs (left) and desires (right).}
        \label{fig:consistency}
\end{figure}

\noindent \textbf{Internal consistency.} The proportion of news with ``unfair'' plans across different belief and desire categories are shown in \figurename~\ref{fig:consistency}. 
Subfigure (a) shows that ``Against'' beliefs have the highest proportion of unfair plans, while ``Neutral'' beliefs align with fairer writing. Subfigure (b) reveals that articles driven by psychological fulfillment exhibit the highest proportion of unfair plans, whereas public interest-driven articles have the lowest proportion.
These findings illustrate the consistency among the internal elements of intent, aligns with the ``consistency criterion'' theory of intent \citep{dennett1987intentional, tomasello2023having}: Intents, as mental states intricately linked to action, drive agents to formulate plans that effectively achieve their desires based on their belief of the situation \citep{audi1993action, heider1958psychology}. 

\begin{figure}[ht]
    \centering
        \includegraphics[width=.75\linewidth]{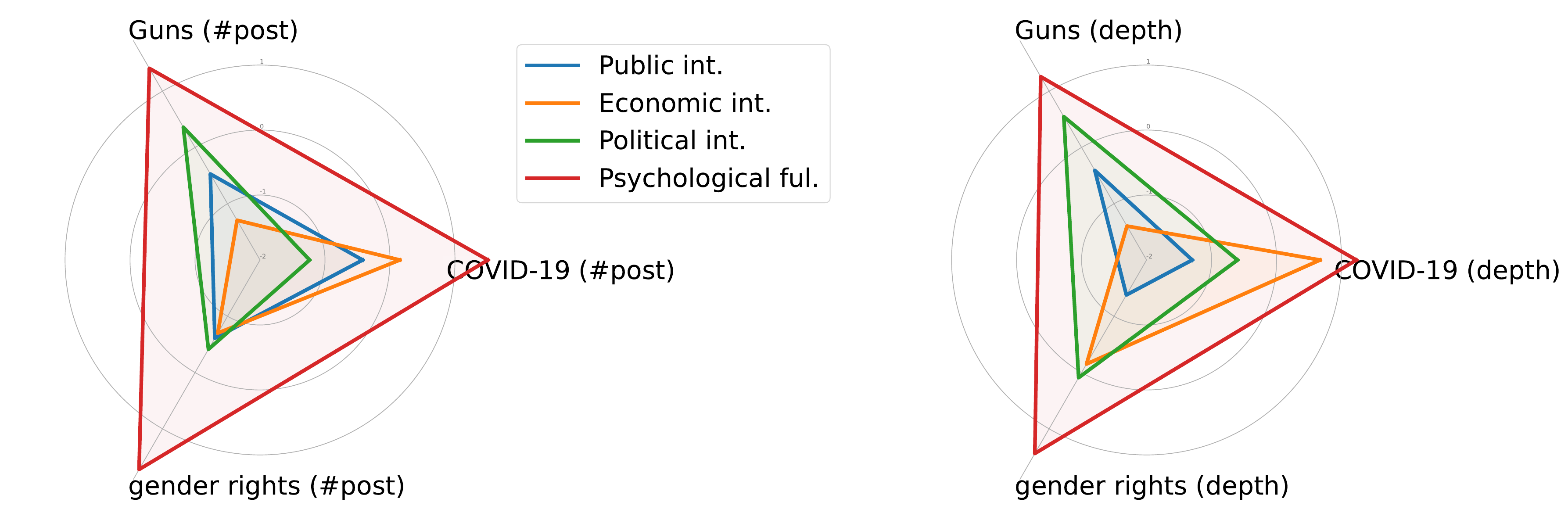}
        \caption{Desire-Outcome connection in average post (left) count and average reply depth (right) over three topics.}
        \label{fig:desire-social}
\end{figure}

\noindent \textbf{External connection.} We selected three topics (COVID-19, gender rights, and gun control) to analyze the desire dimension in news intent and its relationship to real-world consequences, as this dimension is most closely related to outcomes. We utilized post count and discussion depth as outcome proxies to statistically assess the relationship between news intent and these outcomes.
The data was standardized using Z-scores to ensure uniform representation across all subjects, $z = \frac{x - \mu}{\sigma}$, where $x$ denotes the original data point, $\mu$ is the mean of the data point, and $\sigma$ is the standard deviation of the set.

As shown in \figureautorefname~\ref{fig:desire-social}, the intent behind news production correlates with social media engagement. Psychological fulfillment (\textcolor[HTML]{d62728}{red lines}) leads to the highest number of posts and discussion depth due to its sensational nature. Economic interests (\textcolor[HTML]{fbcb57}{yellow lines}) stimulate more discussions on COVID-19, potentially driven by conspiracy theories and the economic repercussions of the pandemic. Political interests (\textcolor[HTML]{2ca02c}{green lines}) provoke in-depth debates, exhibiting greater relative discussion depth despite fewer overall discussions, especially on gender rights. Public interest news (\textcolor[HTML]{2f7eb6}{blue lines}), often leaning toward educational or investigative journalism, elicits less reaction, except for gun-related topics, which are highly contentious.
These findings highlight that modeling news production as an intentional action offers deeper insights into the news lifecycle: it links intents, actions, and consequences, providing a comprehensive framework for future research.

\subsection{Further Analysis}

\begin{table*}[htbp]
\small
\centering
\renewcommand \arraystretch{1.1}
\setlength\tabcolsep{3pt}
\caption{Ablation study of each intent element and its impact on the best-performing model across three application tasks. The {$\Delta$} (marked in \colorbox{verylightgray}{gray}) values indicate relative performance changes compared to using full \ModelName.}
\label{tab:ablation_application}
    \begin{minipage}[htbp]{0.329\textwidth}
    {\small
    \centering
    \begin{tabular}{
      c@{\hspace{0.4cm}}
      c@{\hspace{0.4cm}}
      >{\columncolor{verylightgray}}c@{\hspace{0.4cm}}
    }
    \toprule
    \multicolumn{3}{c}{ {Fake news detection}} \\
    \specialrule{0.04em}{0.2pt}{0.4pt}
    MSynFD & F1$_{\text{fake}}\uparrow$ & \cellcolor{verylightgray}{$\Delta$} \\
    \specialrule{0em}{0.3pt}{0.3pt}
    \hline
    \specialrule{0em}{0.3pt}{0.3pt}
    \hline
    \specialrule{0em}{1pt}{1pt}
    \small
    + \ModelName & \textbf{.725} & - \\
    \; w/o $\mathcal{B}$ & .716 & $-$1.24\% \\
    \; w/o $\mathcal{D}$ & .719 & $-$0.83\% \\
    \; w/o $\mathcal{P}$ & .720 & $-$0.69\% \\
    \bottomrule
    \end{tabular}
    }
    \end{minipage}
    \begin{minipage}[t]{0.329\textwidth}
    \centering
    {\small
    \begin{tabular}{
      c@{\hspace{0.4cm}}
      c@{\hspace{0.4cm}}
      >{\columncolor{verylightgray}}c@{\hspace{0.4cm}}
    }
    \toprule
    \multicolumn{3}{c}{ {News popularity detection}} \\
    \specialrule{0.04em}{0.2pt}{0.4pt}
    F$^4$ & RMSE$\downarrow$ & \cellcolor{verylightgray}{$\Delta$} \\
    \specialrule{0em}{0.3pt}{0.3pt}
    \hline
    \specialrule{0em}{0.3pt}{0.3pt}
    \hline
    \specialrule{0em}{1pt}{1pt}
    + \ModelName & \textbf{.018} & - \\
    \; w/o $\mathcal{B}$ & .020 & $-$1.11\% \\
    \; w/o $\mathcal{D}$ & .023 & $-$2.78\% \\
    \; w/o $\mathcal{P}$ & .022 & $-$2.22\% \\
    \bottomrule
    \end{tabular}
    }
    \end{minipage}
    \begin{minipage}[t]{0.329\textwidth}
    \centering
    {\small
    \begin{tabular}{
      c@{\hspace{0.4cm}}
      c@{\hspace{0.4cm}}
      >{\columncolor{verylightgray}}c@{\hspace{0.4cm}}
    }
    \toprule
    \multicolumn{3}{c}{ {Propaganda detection}} \\
    \specialrule{0.04em}{0.2pt}{0.4pt}
    DET & macF1$\uparrow$ & \cellcolor{verylightgray}{$\Delta$} \\
    \specialrule{0em}{0.3pt}{0.3pt}
    \hline
    \specialrule{0em}{0.3pt}{0.3pt}
    \hline
    \specialrule{0em}{1pt}{1pt}
    + \ModelName & \textbf{.944} & - \\
    \; w/o $\mathcal{B}$ & .939 & $-$0.53\% \\
    \; w/o $\mathcal{D}$ & .941 & $-$0.32\% \\
    \; w/o $\mathcal{P}$ & .938 & $-$0.64\% \\
    \bottomrule
    \end{tabular}
    }
    \end{minipage}
\hfill
\end{table*}

To gain a deeper understanding of the behavior of each intent component—belief, desire, and plan—across application tasks, we conducted an in-depth investigation into how each intent component affects performance across different application scenarios when utilizing our proposed \ModelName model. 
Specifically, we include:
(1) an ablation study on the contribution of each intent element across three application tasks, and
(2) a token-level attention visualization, which illustrates how each intent component attends to different parts of the input text across these tasks.

\noindent\textbf{Ablation study of each intent element across three application tasks.}  
The ablation study is conducted by selectively removing each intent component—belief, desire, or plan—from the integrated \ModelName model. The intent features generated by these model variants are subsequently utilized to assist the best-performing model in application tasks. The averaged results over 5 runs are shown in Table~\ref{tab:ablation_application}.  The results show that removing any single component consistently leads to a drop in performance, indicating that each intent element contributes unique and essential information. These findings empirically confirm the distinct functional roles of belief, desire, and plan in modeling news intent. 
For instance, the removal of the belief component significantly degrades performance in fake news detection, whereas the desire component plays a more critical role in predicting news popularity, likely due to its sensitivity to extravagant language and emotional appeal.

\begin{figure}[htbp]
    \centering
    \begin{subfigure}[b]{\linewidth}
        \centering
        \includegraphics[width=\linewidth]{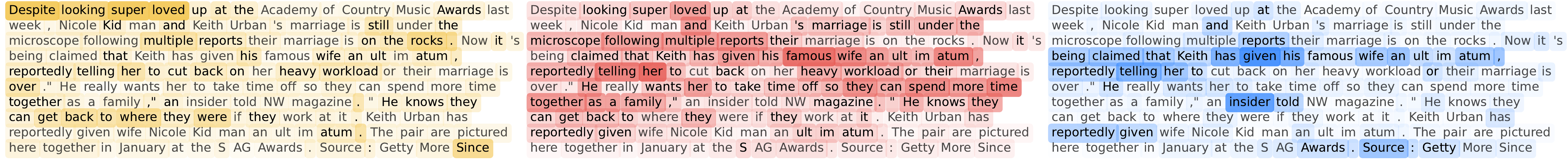}
        \caption{Visualization for fake news detection.}
    \end{subfigure}
    \begin{subfigure}[b]{\linewidth}
        \centering
        \includegraphics[width=\linewidth]{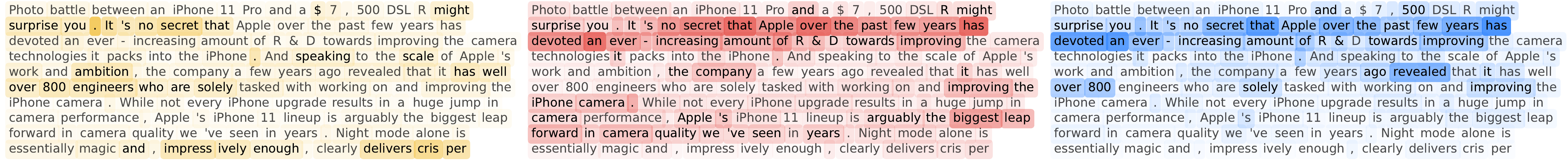}
        \caption{Visualization for news popularity prediction.}
    \end{subfigure}
    \begin{subfigure}[b]{\linewidth}
        \centering
        \includegraphics[width=\linewidth]{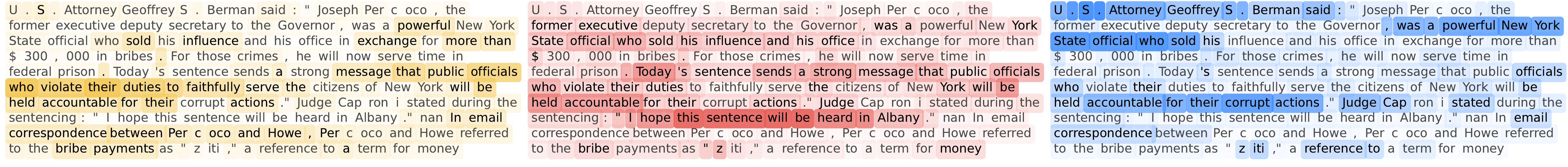}
        \caption{Visualization for propaganda detection.}
    \end{subfigure}
    \caption{
        Token-level gradient-based visualizations of intent components --- belief, desire, and plan --- for three application tasks: (a) fake news detection, (b) news popularity prediction, and (c) propaganda detection. Token background colors encode attention scores over intent components, where \textcolor{yellowv}{\textbf{yellow}} represents \textit{belief}, \textcolor{redv}{\textbf{red}} represents \textit{desire}, and \textcolor{bluev}{\textbf{blue}} represents \textit{plan}, with darker shades representing higher values.
        }
    \label{fig:attention_vis}
\end{figure}

\noindent\textbf{Visualization of each intent component's attention over the input text.}  We compute token-level gradient-based attribution scores from the corresponding expert modules in our \ModelName model. Specifically, for the input news instance, we perform a forward pass through the model to extract the belief, desire, and plan features, then compute the gradient of each feature with respect to the hidden states of the input tokens. The magnitude of these gradients reflects the sensitivity of each intent component to each token, effectively indicating the varying levels of attention that each intent component pays to the input.

As shown in \figurename~\ref{fig:attention_vis}, each visualization example demonstrates how the belief, desire, and plan components focus on different aspects of the input depending on the task. Consistent with the results presented in Table 10, these components provide complementary perspectives to the application tasks, with each playing a distinct role; the absence of any one of these components leads to a decline in performance.
In fake news detection, the model assigns high importance to emotionally charged phrases and speculative language (e.g., ``given his famous wife an ultimatum'', ``reportedly telling'', ``insider told''), particularly through the \textit{desire} and \textit{plan} channels, thereby aiding in the identification of potential patterns in fake news. For news popularity prediction, the \textit{belief} and \textit{desire} components emphasize attitude (e.g., ``impressively enough'', ``devoted an ever-increasing''), while \textit{plan} focuses on factual details (e.g., ``over 800 engineers''). The strong expressions of attitude captured by the model contribute positively to predicting the popularity of the news. In propaganda detection, \textit{belief} and \textit{desire} tend to highlight public accountability statements and value-laden expressions (e.g., ``will be heard'', ``violate their duties''), with \textit{plan} reflecting the persuasive strategy, such as direct quotes (``U.S. Attorney Geoffrey S. Berman said'').

\subsection{Limitations and Future Work}

Despite our proposed approach offering a theoretically grounded and practically effective framework for intent modeling in news articles, there is still space for future directions.
Currently, our method is designed and evaluated solely on news with unimodal textual format, which limits its applicability in multimodal or interactive formats increasingly common in modern journalism.
Expanding news intent research to encompass such richer modalities is needed to build a more comprehensive analysis system.
Future study, such as constructing multimodal intent datasets, is required to establish the baseline and benchmark for more news formats.
Besides, it also requires careful extraction and alignment of cross-modal signals (e.g., image-caption pairs, video transcripts) via cross-modal interaction designs, such as integrating our text-based framework with vision-language pretraining models such as CLIP \citep{radford2021learning, zhang2024vision}, and cross-modal transformers \citep{xu2023multimodal}. 
The investigation of these problems is left for future work.

\section{Related Work} \label{sec:background}

\subsection{General Intent Understanding}

Recent advances in intent understanding span multiple domains: Dialogue systems focus on decoding user goals through utterance patterns (e.g., service requests \citep{louvan2020recent}), while recommendation systems infer user intents from behavioral traces (e.g., purchase propensity \citep{chen2022intent, liu2024learning}). These approaches typically operationalize intent as \textit{explicit functional objectives} directly tied to observable interactions. 

Our work, however, addresses the unique challenge of news intent. News intent fundamentally diverges by operating within \textit{asymmetric persuasion contexts}. Unlike conversational goals that align with participants' mutual awareness \citep{schwarz1991base}, news narratives often conceal their core intents beneath surface-level informativeness \citep{schwarz2014cognition}. Thus, a theoretically grounded framework is essential to navigate this complexity and to unravel the complex interplay between the inner components of intent, news production practices, and their broader social impact.

\subsection{Perception of News Intent}

Prior work related to news intent mining falls into two paradigms with inherent limitations:
Empirical taxonomy approaches \citep{kang2019health, dharawat2022drink} manually categorize intents within narrow domains (e.g., health issues) using topic-specific labels \citep{caballero2023exploring}. While yielding fine-grained classes (e.g., Electoral legitimacy concerns), these taxonomies lack transferable schemas; for instance, a model trained on taxonomies specific to Korean political news \citep{kim2023new} fails to capture intents beyond that topic.
Heuristic feature engineering studies \citep{chen2020sirenless, wang2024misinformation} approximate intent through surface correlates, such as lexical markers and sentiment shifts. These handcrafted or unsupervised proxies treat intent as a variant of stylistic features, focusing on feature selection or optimization for a single intent category (i.e., deception intent). As a result, they are unable to model diverse intents—a critical drawback considering the multi-intent compositions present in news narratives (e.g., simultaneous deception and agenda-setting).
\citet{gabriel2022misinfo} proposed a vague concept of intent, relying on annotators' intuition and free-text descriptions, which hampers precise assessment and efficient usage.
These attempts are limited to specific topics and datasets, lacking broader theoretical foundations. 

In contrast, we propose the \FrameName framework, grounded in interdisciplinary studies, to eliminate the ambiguity of intent and provide a structured framework for intent modeling and practical application, enabling cross-task applications.

\section{Conclusion} \label{sec:conclusion}

This research introduces a new perspective in news studies by conceptualizing news production as intentional action and offering a deconstructive analysis of news intent. We present \FrameName, a conceptual deconstruction-based news intent understanding framework grounded in interdisciplinary studies, to systematically investigate news intent. Then, we enhanced the perception capabilities of LLMs regarding news intent and compiled a dataset containing 142.5k fine-grained labels. Finally, we validated the practical value of news intent and our proposed computational modeling approach in downstream news-related tasks. Our findings demonstrate the internal consistency and external connection of news intent, providing valuable insights for news intent cognition and computational social science.

\normalem

\begin{thebibliography}{102}
\providecommand{\natexlab}[1]{#1}

\bibitem[{Ali and Hassan(2022)}]{ali2022survey}
Mohammad Ali and Naeemul Hassan. 2022.
\newblock A survey of computational framing analysis approaches.
\newblock In \emph{Proceedings of the 2022 Conference on Empirical Methods in Natural Language Processing}, pages 9335--9348.

\bibitem[{Ames and Fiske(2015)}]{ames2015perceived}
Daniel~L Ames and Susan~T Fiske. 2015.
\newblock Perceived intent motivates people to magnify observed harms.
\newblock \emph{Proceedings of the National Academy of Sciences}, 112(12):3599--3605.

\bibitem[{Anscombe(2000)}]{anscombe2000intention}
Gertrude Elizabeth~Margaret Anscombe. 2000.
\newblock \emph{Intention}.
\newblock Harvard University Press.

\bibitem[{Armstrong(1973)}]{armstrong1973belief}
David~Malet Armstrong. 1973.
\newblock \emph{Belief, truth and knowledge}.
\newblock Cambridge University Press.

\bibitem[{Asp(2007)}]{asp2007fairness}
Kent Asp. 2007.
\newblock Fairness, Informativeness and Scrutiny: The Role of News Media in Democracy.
\newblock \emph{Nordicom Review}, 28.

\bibitem[{Audi(1993)}]{audi1993action}
Robert Audi. 1993.
\newblock \emph{Action, intention, and reason}.
\newblock Cornell University Press.

\bibitem[{Barr{\'o}n-Cedeno et~al.(2019)Barr{\'o}n-Cedeno, Jaradat, Da~San~Martino, and Nakov}]{barron2019proppy}
Alberto Barr{\'o}n-Cedeno, Israa Jaradat, Giovanni Da~San~Martino, and Preslav Nakov. 2019.
\newblock Proppy: Organizing the news based on their propagandistic content.
\newblock \emph{Information Processing \& Management}, 56(5):1849--1864.

\bibitem[{Boichak(2021)}]{boichak2020digital}
Olga Boichak. 2021.
\newblock \href {https://doi.org/10.1093/oxfordhb/9780197510636.013.31} {{Digital War: Mediatized Conflicts in Sociological Perspective}}.
\newblock In \emph{{The Oxford Handbook of Digital Media Sociology}}, pages 511--527. Oxford University Press.

\bibitem[{Bratman(1984)}]{bratman1984two}
Michael Bratman. 1984.
\newblock Two faces of intention.
\newblock \emph{The Philosophical Review}, 93(3):375--405.

\bibitem[{Bratman(1987)}]{bratman1987intention}
Michael Bratman. 1987.
\newblock \emph{Intention, plans, and practical reason}.
\newblock Number~10 in The David Hume Series. Harvard University Press, Cambridge, MA.

\bibitem[{Bratman(2009)}]{bratman2009intention}
Michael~E Bratman. 2009.
\newblock Intention, practical rationality, and self-governance.
\newblock \emph{Ethics}, 119(3):411--443.

\bibitem[{Brown et~al.(2020)Brown, Mann, Ryder, Subbiah, Kaplan, Dhariwal, Neelakantan, Shyam, Sastry, Askell et~al.}]{brown2020language}
Tom Brown, Benjamin Mann, Nick Ryder, Melanie Subbiah, Jared~D Kaplan, Prafulla Dhariwal, Arvind Neelakantan, Pranav Shyam, Girish Sastry, Amanda Askell, et~al. 2020.
\newblock Language Models are Few-Shot Learners.
\newblock \emph{Advances in Neural Information Processing Systems}, 33:1877--1901.

\bibitem[{Caballero~Hinojosa(2023)}]{caballero2023exploring}
Alberto Caballero~Hinojosa. 2023.
\newblock Exploring the Power of Large Language Models: News Intention Detection using Adaptive Learning Prompting.
\newblock \emph{master thesis}.

\bibitem[{Chen et~al.(2020)Chen, Lo, and Qu}]{chen2020sirenless}
Xumeng Chen, Leo Yu-Ho Lo, and Huamin Qu. 2020.
\newblock SirenLess: Reveal the intention behind news.
\newblock \emph{arXiv preprint arXiv:2001.02731}.

\bibitem[{Chen et~al.(2022)Chen, Liu, Li, McAuley, and Xiong}]{chen2022intent}
Yongjun Chen, Zhiwei Liu, Jia Li, Julian McAuley, and Caiming Xiong. 2022.
\newblock Intent contrastive learning for sequential recommendation.
\newblock In \emph{Proceedings of the ACM Web Conference 2022}, pages 2172--2182.

\bibitem[{Chernyavskiy et~al.(2024)Chernyavskiy, Ilvovsky, and Nakov}]{chernyavskiy2024unleashing}
Alexander Chernyavskiy, Dmitry Ilvovsky, and Preslav Nakov. 2024.
\newblock Unleashing the Power of Discourse-Enhanced Transformers for Propaganda Detection.
\newblock In \emph{Proceedings of the 18th Conference of the European Chapter of the Association for Computational Linguistics (Volume 1: Long Papers)}, pages 1452--1462.

\bibitem[{Chung et~al.(2024)Chung, Hou, Longpre, Zoph, Tay, Fedus, Li, Wang, Dehghani, Brahma et~al.}]{chung2024scaling}
Hyung~Won Chung, Le~Hou, Shayne Longpre, Barret Zoph, Yi~Tay, William Fedus, Yunxuan Li, Xuezhi Wang, Mostafa Dehghani, Siddhartha Brahma, et~al. 2024.
\newblock Scaling instruction-finetuned language models.
\newblock \emph{Journal of Machine Learning Research}, 25(70):1--53.

\bibitem[{Cushman(2015)}]{cushman2015deconstructing}
Fiery Cushman. 2015.
\newblock Deconstructing intent to reconstruct morality.
\newblock \emph{Current Opinion in Psychology}, 6:97--103.

\bibitem[{Cushman and Mele(2008)}]{cushman2008intentional}
Fiery Cushman and Alfred Mele. 2008.
\newblock Intentional action.
\newblock \emph{Experimental philosophy}, 1:171.

\bibitem[{Dennett(1987)}]{dennett1987intentional}
Daniel~Clement Dennett. 1987.
\newblock \emph{The intentional stance}.
\newblock MIT press.

\bibitem[{Dharawat et~al.(2022)Dharawat, Lourentzou, Morales, and Zhai}]{dharawat2022drink}
Arkin Dharawat, Ismini Lourentzou, Alex Morales, and ChengXiang Zhai. 2022.
\newblock Drink bleach or do what now? covid-hera: A study of risk-informed health decision making in the presence of covid-19 misinformation.
\newblock In \emph{Proceedings of the International AAAI Conference on Web and Social Media}, volume~16, pages 1218--1227.

\bibitem[{Fan et~al.(2021)Fan, Lin, Li, and Zou}]{fan2021news}
Shuai Fan, Chen Lin, Hui Li, and Quan Zou. 2021.
\newblock News popularity prediction with local-global long-short-term embedding.
\newblock In \emph{International Conference on Web Information Systems Engineering}, pages 79--93. Springer.

\bibitem[{Fleiss(1971)}]{fleiss1971measuring}
Joseph~L Fleiss. 1971.
\newblock Measuring nominal scale agreement among many raters.
\newblock \emph{Psychological bulletin}, 76(5):378.

\bibitem[{Gabriel et~al.(2022)Gabriel, Hallinan, Sap, Nguyen, Roesner, Choi, and Choi}]{gabriel2022misinfo}
Saadia Gabriel, Skyler Hallinan, Maarten Sap, Pemi Nguyen, Franziska Roesner, Eunsol Choi, and Yejin Choi. 2022.
\newblock Misinfo Reaction Frames: Reasoning about Readers’ Reactions to News Headlines.
\newblock In \emph{Proceedings of the 60th Annual Meeting of the Association for Computational Linguistics (Volume 1: Long Papers)}, pages 3108--3127.

\bibitem[{Gergely and Csibra(2003)}]{gergely2003teleological}
Gy{\"o}rgy Gergely and Gergely Csibra. 2003.
\newblock Teleological reasoning in infancy: The na{\i}ve theory of rational action.
\newblock \emph{Trends in cognitive sciences}, 7(7):287--292.

\bibitem[{Gilardi et~al.(2023)Gilardi, Alizadeh, and Kubli}]{Gilardi2023ChatGPTOC}
Fabrizio Gilardi, Meysam Alizadeh, and Ma{\"e}l Kubli. 2023.
\newblock ChatGPT outperforms crowd workers for text-annotation tasks.
\newblock \emph{Proceedings of the National Academy of Sciences of the United States of America}, 120.

\bibitem[{Hasanain et~al.(2024)Hasanain, Ahmad, and Alam}]{hasanain2024can}
Maram Hasanain, Fatema Ahmad, and Firoj Alam. 2024.
\newblock Can GPT-4 Identify Propaganda? Annotation and Detection of Propaganda Spans in News Articles.
\newblock In \emph{Proceedings of the 2024 Joint International Conference on Computational Linguistics, Language Resources and Evaluation (LREC-COLING 2024)}, pages 2724--2744.

\bibitem[{Heider(2013)}]{heider1958psychology}
Fritz Heider. 2013.
\newblock \emph{The psychology of interpersonal relations}.
\newblock Psychology Press.

\bibitem[{Hu et~al.(2024)Hu, Sheng, Cao, Shi, Li, Wang, and Qi}]{hu2024bad}
Beizhe Hu, Qiang Sheng, Juan Cao, Yuhui Shi, Yang Li, Danding Wang, and Peng Qi. 2024.
\newblock Bad actor, good advisor: Exploring the role of large language models in fake news detection.
\newblock In \emph{Proceedings of the AAAI Conference on Artificial Intelligence}, volume~38, pages 22105--22113.

\bibitem[{Hu et~al.(2023)Hu, Sheng, Cao, Zhu, Wang, Wang, and Jin}]{hu2023learn}
Beizhe Hu, Qiang Sheng, Juan Cao, Yongchun Zhu, Danding Wang, Zhengjia Wang, and Zhiwei Jin. 2023.
\newblock Learn over Past, Evolve for Future: Forecasting Temporal Trends for Fake News Detection.
\newblock In \emph{Proceedings of the 61st Annual Meeting of the Association for Computational Linguistics (Volume 5: Industry Track)}, pages 116--125.

\bibitem[{Immorlica et~al.(2024)Immorlica, Jagadeesan, and Lucier}]{immorlica2024clickbait}
Nicole Immorlica, Meena Jagadeesan, and Brendan Lucier. 2024.
\newblock Clickbait vs. quality: How engagement-based optimization shapes the content landscape in online platforms.
\newblock In \emph{Proceedings of the ACM on Web Conference 2024}, pages 36--45.

\bibitem[{Ireton and Posetti(2018)}]{ireton2018journalism}
Cherilyn Ireton and Julie Posetti. 2018.
\newblock \emph{Journalism, fake news \& disinformation: handbook for journalism education and training}.
\newblock Unesco Publishing.

\bibitem[{Jowett and O'donnell(2018)}]{jowett2018propaganda}
Garth~S Jowett and Victoria O'donnell. 2018.
\newblock \emph{Propaganda \& persuasion}.
\newblock Sage publications.

\bibitem[{Kang(2019)}]{kang2019health}
EunKyo Kang. 2019.
\newblock Health information in the news media: Evaluating sources and subject of articles, and the intention to advertise.
\newblock In \emph{Proceedings of the 9th International Conference on Digital Public Health}, pages 121--121.

\bibitem[{Kim et~al.(2023)Kim, Lee, and Na}]{kim2023new}
Beomjune Kim, Eunsun Lee, and Dongbin Na. 2023.
\newblock A New Korean Text Classification Benchmark for Recognizing the Political Intents in Online Newspapers.
\newblock \emph{arXiv preprint arXiv:2311.01712}.

\bibitem[{Kingma and Ba(2015)}]{kingma2015adam}
Diederik~P Kingma and Jimmy Ba. 2015.
\newblock Adam: A Method for Stochastic Optimization.
\newblock In \emph{Proceedings of the 3rd International Conference Learning Representations (Poster)}, pages 1--15.

\bibitem[{K{\"u}{\c{c}}{\"u}k and Can(2020)}]{kuccuk2020stance}
Dilek K{\"u}{\c{c}}{\"u}k and Fazli Can. 2020.
\newblock Stance detection: A survey.
\newblock \emph{ACM Computing Surveys}, 53(1):1--37.

\bibitem[{Landis and Koch(1977)}]{landis1977measurement}
J~Richard Landis and Gary~G Koch. 1977.
\newblock The measurement of observer agreement for categorical data.
\newblock \emph{biometrics}, pages 159--174.

\bibitem[{Liu et~al.(2024)Liu, Zhou, Song, Ouyang, Li, Jing, Yu, and Ng}]{liu2024learning}
Huafeng Liu, Mingjie Zhou, Mingyang Song, Deqiang Ouyang, Yawen Li, Liping Jing, Jian Yu, and Michael~K Ng. 2024.
\newblock Learning hierarchical preferences for recommendation with mixture intention neural stochastic processes.
\newblock \emph{IEEE Transactions on Knowledge and Data Engineering}.

\bibitem[{Liu et~al.(2019)Liu, Ott, Goyal, Du, Joshi, Chen, Levy, Lewis, Zettlemoyer, and Stoyanov}]{liu2019roberta}
Yinhan Liu, Myle Ott, Naman Goyal, Jingfei Du, Mandar Joshi, Danqi Chen, Omer Levy, Mike Lewis, Luke Zettlemoyer, and Veselin Stoyanov. 2019.
\newblock {RoBERTa}: A robustly optimized bert pretraining approach.
\newblock \emph{arXiv preprint arXiv:1907.11692}.

\bibitem[{Louvan and Magnini(2020)}]{louvan2020recent}
Samuel Louvan and Bernardo Magnini. 2020.
\newblock Recent Neural Methods on Slot Filling and Intent Classification for Task-Oriented Dialogue Systems: A Survey.
\newblock In \emph{Proceedings of the 28th International Conference on Computational Linguistics}, pages 480--496.

\bibitem[{Lu et~al.(2023)Lu, Xu, Zhang, Min, Yang, and Lin}]{lu2023facilitating}
Junyu Lu, Bo~Xu, Xiaokun Zhang, Changrong Min, Liang Yang, and Hongfei Lin. 2023.
\newblock Facilitating Fine-grained Detection of Chinese Toxic Language: Hierarchical Taxonomy, Resources, and Benchmarks.
\newblock In \emph{Proceedings of the 61st Annual Meeting of the Association for Computational Linguistics (Volume 1: Long Papers)}, pages 16235--16250.

\bibitem[{Luvembe et~al.(2023)Luvembe, Li, Li, Liu, and Xu}]{luvembe2023dual}
Alex~Munyole Luvembe, Weimin Li, Shaohua Li, Fangfang Liu, and Guiqiong Xu. 2023.
\newblock Dual emotion based fake news detection: A deep attention-weight update approach.
\newblock \emph{Information Processing \& Management}, 60(4):103354.

\bibitem[{Mahony and Chen(2024)}]{mahony2024concerns}
Simon Mahony and Qing Chen. 2024.
\newblock Concerns about the role of artificial intelligence in journalism, and media manipulation.
\newblock \emph{Journalism}, page 14648849241263293.

\bibitem[{Malle and Knobe(1997)}]{malle1997folk}
Bertram~F Malle and Joshua Knobe. 1997.
\newblock The folk concept of intentionality.
\newblock \emph{Journal of experimental social psychology}, 33(2):101--121.

\bibitem[{Mele(1997)}]{mele1997philosophy}
Alfred~R Mele. 1997.
\newblock The philosophy of action.
\newblock \emph{Donald Davidson}, page~64.

\bibitem[{Mendoza et~al.(2020)Mendoza, Parra, and Soto}]{mendoza2020gene}
Marcelo Mendoza, Denis Parra, and {\'A}lvaro Soto. 2020.
\newblock GENE: Graph generation conditioned on named entities for polarity and controversy detection in social media.
\newblock \emph{Information Processing \& Management}, 57(6):102366.

\bibitem[{Mittal et~al.(2024)Mittal, Chowdhury, Guhan, Chelluri, and Manocha}]{mittal2024towards}
Trisha Mittal, Sanjoy Chowdhury, Pooja Guhan, Snikitha Chelluri, and Dinesh Manocha. 2024.
\newblock Towards determining perceived audience intent for multimodal social media posts using the theory of reasoned action.
\newblock \emph{Scientific Reports}, 14(1):10606.

\bibitem[{Murayama et~al.(2022)Murayama, Hisada, Uehara, Wakamiya, and Aramaki}]{murayama2022annotation}
Taichi Murayama, Shohei Hisada, Makoto Uehara, Shoko Wakamiya, and Eiji Aramaki. 2022.
\newblock Annotation-Scheme Reconstruction for “Fake News” and Japanese Fake News Dataset.
\newblock In \emph{Proceedings of the Thirteenth Language Resources and Evaluation Conference}, pages 7226--7234.

\bibitem[{Nan et~al.(2024)Nan, Sheng, Cao, Hu, Wang, and Li}]{nan2024let}
Qiong Nan, Qiang Sheng, Juan Cao, Beizhe Hu, Danding Wang, and Jintao Li. 2024.
\newblock Let silence speak: Enhancing fake news detection with generated comments from large language models.
\newblock In \emph{Proceedings of the 33rd ACM International Conference on Information and Knowledge Management}, pages 1732--1742.

\bibitem[{Nan et~al.(2025)Nan, Sheng, Cao, Zhu, Wang, Yang, and Li}]{nan2025exploiting}
Qiong Nan, Qiang Sheng, Juan Cao, Yongchun Zhu, Danding Wang, Guang Yang, and Jintao Li. 2025.
\newblock Exploiting user comments for early detection of fake news prior to users’ commenting.
\newblock \emph{Frontiers of Computer Science}, 19(10):1910354.

\bibitem[{Nelson et~al.(2024)Nelson, Lewis, and Cowley}]{nelson2024money}
Jacob~L Nelson, Seth~C Lewis, and Brent Cowley. 2024.
\newblock ‘Money is the root of all evil.’How the business of journalism shapes trust in news.
\newblock \emph{Journalism}, page 14648849241246929.

\bibitem[{Newcomb(1958)}]{newcomb1958psychology}
Theodore Newcomb. 1958.
\newblock The Psychology of Interpersonal Relations.

\bibitem[{OpenAI(2022)}]{chatgpt}
OpenAI. 2022.
\newblock {ChatGPT}: Optimizing Language Models for Dialogue.
\newblock \url{https://openai.com/blog/chatgpt/}.
\newblock Accessed: 2023-08-13.

\bibitem[{Pettit and Smith(1996)}]{pettit1996freedom}
Philip Pettit and Michael Smith. 1996.
\newblock Freedom in belief and desire.
\newblock \emph{The Journal of Philosophy}, 93(9):429--449.

\bibitem[{Phillips et~al.(2002)Phillips, Wellman, and Spelke}]{phillips2002infants}
Ann~T Phillips, Henry~M Wellman, and Elizabeth~S Spelke. 2002.
\newblock Infants' ability to connect gaze and emotional expression to intentional action.
\newblock \emph{Cognition}, 85(1):53--78.

\bibitem[{Piskorski et~al.(2023)Piskorski, Stefanovitch, Da~San~Martino, and Nakov}]{piskorski2023semeval}
Jakub Piskorski, Nicolas Stefanovitch, Giovanni Da~San~Martino, and Preslav Nakov. 2023.
\newblock Semeval-2023 task 3: Detecting the category, the framing, and the persuasion techniques in online news in a multi-lingual setup.
\newblock In \emph{Proceedings of the 17th International Workshop on Semantic Evaluation (SemEval-2023)}, pages 2343--2361.

\bibitem[{Porlezza(2024)}]{porlezza2024datafication}
Colin Porlezza. 2024.
\newblock The datafication of digital journalism: A history of everlasting challenges between ethical issues and regulation.
\newblock \emph{Journalism}, 25(5):1167--1185.

\bibitem[{Pramanick et~al.(2021)Pramanick, Sharma, Dimitrov, Akhtar, Nakov, and Chakraborty}]{pramanick2021momenta}
Shraman Pramanick, Shivam Sharma, Dimitar Dimitrov, Md~Shad Akhtar, Preslav Nakov, and Tanmoy Chakraborty. 2021.
\newblock MOMENTA: A Multimodal Framework for Detecting Harmful Memes and Their Targets.
\newblock In \emph{Findings of the Association for Computational Linguistics: EMNLP 2021}, pages 4439--4455.

\bibitem[{Quandt(2024)}]{quandt2024euphoria}
Thorsten Quandt. 2024.
\newblock Euphoria, disillusionment and fear: Twenty-five years of digital journalism.
\newblock \emph{Journalism}, 25(5):1186--1203.

\bibitem[{Qwen-Team(2025)}]{qwq32b}
Qwen-Team. 2025.
\newblock \href {https://qwenlm.github.io/blog/qwq-32b/} {QwQ-32B: Embracing the Power of Reinforcement Learning}.

\bibitem[{Radford et~al.(2021)Radford, Kim, Hallacy, Ramesh, Goh, Agarwal, Sastry, Askell, Mishkin, Clark et~al.}]{radford2021learning}
Alec Radford, Jong~Wook Kim, Chris Hallacy, Aditya Ramesh, Gabriel Goh, Sandhini Agarwal, Girish Sastry, Amanda Askell, Pamela Mishkin, Jack Clark, et~al. 2021.
\newblock Learning transferable visual models from natural language supervision.
\newblock In \emph{International conference on machine learning}, pages 8748--8763. PmLR.

\bibitem[{Randolph(2005)}]{Randolph2005FreeMarginalMK}
Justus~J. Randolph. 2005.
\newblock \href {https://eric.ed.gov/?id=ED490661} {Free-Marginal Multirater Kappa (multirater K[free]): An Alternative to Fleiss' Fixed-Marginal Multirater Kappa.}
\newblock In \emph{Joensuu Learning and Instruction Symposium}, pages 1--20.

\bibitem[{Sajwani et~al.(2024)Sajwani, El~Setohy, Mekky, Turmakhan, Hassan, El~Zeftawy, El~Herraoui, Afzal, Liao, Mahmoud et~al.}]{sajwani2024frappe}
Ahmed Sajwani, Alaa El~Setohy, Ali Mekky, Diana Turmakhan, Lara Hassan, Mohamed El~Zeftawy, Omar El~Herraoui, Osama Afzal, Qisheng Liao, Tarek Mahmoud, et~al. 2024.
\newblock FRAPPE: FRAming, Persuasion, and Propaganda Explorer.
\newblock In \emph{Proceedings of the 18th Conference of the European Chapter of the Association for Computational Linguistics: System Demonstrations}, pages 207--213.

\bibitem[{Sakketou et~al.(2022)Sakketou, Plepi, Cervero, Geiss, Rosso, and Flek}]{sakketou2022factoid}
Flora Sakketou, Joan Plepi, Riccardo Cervero, Henri~Jacques Geiss, Paolo Rosso, and Lucie Flek. 2022.
\newblock \href {https://aclanthology.org/2022.lrec-1.345/} {{FACTOID}: A New Dataset for Identifying Misinformation Spreaders and Political Bias}.
\newblock In \emph{Proceedings of the Thirteenth Language Resources and Evaluation Conference}, pages 3231--3241, Marseille, France. European Language Resources Association.

\bibitem[{Sassenberg and Winter(2024)}]{sassenberg2024intraindividual}
Kai Sassenberg and Kevin Winter. 2024.
\newblock Intraindividual conflicts reduce the polarization of attitudes.
\newblock \emph{Current Directions in Psychological Science}, 33(3):190--197.

\bibitem[{Schwarz(2014)}]{schwarz2014cognition}
Norbert Schwarz. 2014.
\newblock \emph{Cognition and communication: Judgmental biases, research methods, and the logic of conversation}.
\newblock Psychology Press.

\bibitem[{Schwarz et~al.(1991)Schwarz, Strack, Hilton, and Naderer}]{schwarz1991base}
Norbert Schwarz, Fritz Strack, Denis Hilton, and Gabi Naderer. 1991.
\newblock Base rates, representativeness, and the logic of conversation: The contextual relevance of “irrelevant” information.
\newblock \emph{Social Cognition}, 9(1):67--84.

\bibitem[{Schwitzgebel(2021)}]{sep-belief}
Eric Schwitzgebel. 2021.
\newblock {Belief}.
\newblock In Edward~N. Zalta, editor, \emph{The {Stanford} Encyclopedia of Philosophy}, {W}inter 2021 edition, pages 1--11. Metaphysics Research Lab, Stanford University.

\bibitem[{Sheng et~al.(2022{\natexlab{a}})Sheng, Cao, Bernard, Shu, Li, and Liu}]{sheng2022characterizing}
Qiang Sheng, Juan Cao, H~Russell Bernard, Kai Shu, Jintao Li, and Huan Liu. 2022{\natexlab{a}}.
\newblock Characterizing multi-domain false news and underlying user effects on Chinese Weibo.
\newblock \emph{Information Processing \& Management}, 59(4):102959.

\bibitem[{Sheng et~al.(2022{\natexlab{b}})Sheng, Cao, Zhang, Li, Wang, and Zhu}]{sheng2022zoom}
Qiang Sheng, Juan Cao, Xueyao Zhang, Rundong Li, Danding Wang, and Yongchun Zhu. 2022{\natexlab{b}}.
\newblock \href {https://doi.org/10.18653/v1/2022.acl-long.311} {Zoom Out and Observe: News Environment Perception for Fake News Detection}.
\newblock In \emph{Proceedings of the 60th Annual Meeting of the Association for Computational Linguistics (Volume 1: Long Papers)}, pages 4543--4556. Association for Computational Linguistics.

\bibitem[{Sheng et~al.(2021)Sheng, Zhang, Cao, and Zhong}]{sheng2021integrating}
Qiang Sheng, Xueyao Zhang, Juan Cao, and Lei Zhong. 2021.
\newblock \href {https://doi.org/10.1145/3459637.3482440} {Integrating Pattern- and Fact-based Fake News Detection via Model Preference Learning}.
\newblock In \emph{Proceedings of the 30th ACM International Conference on Information \& Knowledge Management}, pages 1640--1650. Association for Computing Machinery.

\bibitem[{Sheng et~al.(2025)Sheng, Zeng, Tang, Liu, and Zhao}]{sheng2025confusing}
Yaqing Sheng, Weixin Zeng, Jiuyang Tang, Lihua Liu, and Xiang Zhao. 2025.
\newblock Confusing negative commonsense knowledge generation with hierarchy modeling and LLM-enhanced filtering.
\newblock \emph{Information Processing \& Management}, 62(3):104060.

\bibitem[{Shu et~al.(2020)Shu, Mahudeswaran, Wang, Lee, and Liu}]{shu2020fakenewsnet}
Kai Shu, Deepak Mahudeswaran, Suhang Wang, Dongwon Lee, and Huan Liu. 2020.
\newblock Fake{N}ews{N}et: A data repository with news content, social context, and spatiotemporal information for studying fake news on social media.
\newblock \emph{Big Data}, 8(3):171--188.

\bibitem[{Sinelnik and Hovy(2024)}]{sinelnik2024narratives}
Antonina Sinelnik and Dirk Hovy. 2024.
\newblock Narratives at Conflict: Computational Analysis of News Framing in Multilingual Disinformation Campaigns.
\newblock In \emph{Proceedings of the 62nd Annual Meeting of the Association for Computational Linguistics (Volume 4: Student Research Workshop)}, pages 225--237.

\bibitem[{Sinhababu(2017)}]{sinhababu2017humean}
Neil Sinhababu. 2017.
\newblock \emph{Humean nature: How desire explains action, thought, and feeling}.
\newblock Oxford University Press.

\bibitem[{Smith(1994)}]{smith1994moral}
Michael Smith. 1994.
\newblock \emph{The moral problem}.
\newblock Blackwell.

\bibitem[{Son and Fielding(2024)}]{son2024teen}
Catherine Son and Victoria Fielding. 2024.
\newblock “Teen Fled Danger into the Arms of Death”: The Political Agenda Setting Effect of Australian News Media Framing of Violence Against Women.
\newblock \emph{Violence against women}, page 10778012241228291.

\bibitem[{Spinde et~al.(2021)Spinde, Rudnitckaia, Mitrovi{\'c}, Hamborg, Granitzer, Gipp, and Donnay}]{spinde2021automated}
Timo Spinde, Lada Rudnitckaia, Jelena Mitrovi{\'c}, Felix Hamborg, Michael Granitzer, Bela Gipp, and Karsten Donnay. 2021.
\newblock Automated identification of bias inducing words in news articles using linguistic and context-oriented features.
\newblock \emph{Information Processing \& Management}, 58(3):102505.

\bibitem[{Sukthankar et~al.(2014)Sukthankar, Geib, Bui, Pynadath, and Goldman}]{sukthankar2014plan}
Gita Sukthankar, Christopher Geib, Hung~Hai Bui, David Pynadath, and Robert~P Goldman. 2014.
\newblock \emph{Plan, activity, and intent recognition: Theory and practice}.
\newblock Newnes.

\bibitem[{Tomasello(2023)}]{tomasello2023having}
Michael Tomasello. 2023.
\newblock Having intentions, understanding intentions, and understanding communicative intentions.
\newblock In \emph{Developing theories of intention}, pages 63--76. Psychology Press.

\bibitem[{Touvron et~al.(2023)Touvron, Lavril, Izacard, Martinet, Lachaux, Lacroix, Rozi{\`e}re, Goyal, Hambro, Azhar et~al.}]{touvron2023llama}
Hugo Touvron, Thibaut Lavril, Gautier Izacard, Xavier Martinet, Marie-Anne Lachaux, Timoth{\'e}e Lacroix, Baptiste Rozi{\`e}re, Naman Goyal, Eric Hambro, Faisal Azhar, et~al. 2023.
\newblock Llama: Open and efficient foundation language models.
\newblock \emph{arXiv preprint arXiv:2302.13971}.

\bibitem[{Van~Dijk(2013)}]{van2013news}
Teun~A Van~Dijk. 2013.
\newblock \emph{News as discourse}.
\newblock Routledge.

\bibitem[{Wang et~al.(2024)Wang, Li, Li, Fu, Pei, and Wang}]{wang2024misinformation}
Bing Wang, Ximing Li, Changchun Li, Bo~Fu, Songwen Pei, and Shengsheng Wang. 2024.
\newblock Why Misinformation is Created? Detecting them by Integrating Intent Features.
\newblock In \emph{Proceedings of the 33rd ACM International Conference on Information and Knowledge Management}, pages 2304--2314.

\bibitem[{Wardle and Derakhshan(2017)}]{wardle2017information}
Claire Wardle and Hossein Derakhshan. 2017.
\newblock \emph{Information disorder: Toward an interdisciplinary framework for research and policymaking}, volume~27.
\newblock Council of Europe Strasbourg.

\bibitem[{Weld et~al.(2022)Weld, Huang, Long, Poon, and Han}]{weld2022survey}
Henry Weld, Xiaoqi Huang, Siqu Long, Josiah Poon, and Soyeon~Caren Han. 2022.
\newblock A survey of joint intent detection and slot filling models in natural language understanding.
\newblock \emph{ACM Computing Surveys}, 55(8):1--38.

\bibitem[{Wellman et~al.(2001)Wellman, Cross, and Watson}]{wellman2001meta}
Henry~M Wellman, David Cross, and Julanne Watson. 2001.
\newblock Meta-analysis of theory-of-mind development: The truth about false belief.
\newblock \emph{Child development}, 72(3):655--684.

\bibitem[{Wu et~al.(2020)Wu, Qiao, Chen, Wu, Qi, Lian, Liu, Xie, Gao, Wu et~al.}]{wu2020mind}
Fangzhao Wu, Ying Qiao, Jiun-Hung Chen, Chuhan Wu, Tao Qi, Jianxun Lian, Danyang Liu, Xing Xie, Jianfeng Gao, Winnie Wu, et~al. 2020.
\newblock Mind: A large-scale dataset for news recommendation.
\newblock In \emph{Proceedings of the 58th Annual Meeting of the Association for Computational Linguistics}, pages 3597--3606.

\bibitem[{Xiao et~al.(2024)Xiao, Zhang, Shi, Wang, Naseem, and Hu}]{xiao2024msynfd}
Liang Xiao, Qi~Zhang, Chongyang Shi, Shoujin Wang, Usman Naseem, and Liang Hu. 2024.
\newblock MSynFD: Multi-hop Syntax aware Fake News Detection.
\newblock In \emph{Proceedings of the ACM on Web Conference 2024}, pages 4128--4137.

\bibitem[{Xu et~al.(2023)Xu, Zhu, and Clifton}]{xu2023multimodal}
Peng Xu, Xiatian Zhu, and David~A Clifton. 2023.
\newblock Multimodal learning with transformers: A survey.
\newblock \emph{IEEE Transactions on Pattern Analysis and Machine Intelligence}, 45(10):12113--12132.

\bibitem[{Yang et~al.(2024)Yang, Yang, Zhang, Hui, Zheng, Yu, Li, Liu, Huang, Wei, Lin, Yang, Tu, Zhang, Yang, Yang, Zhou, Lin, Dang, Lu, Bao, Yang, Yu, Li, Xue, Zhang, Zhu, Men, Lin, Li, Tang, Xia, Ren, Ren, Fan, Su, Zhang, Wan, Liu, Cui, Zhang, and Qiu}]{qwen2.5}
An~Yang, Baosong Yang, Beichen Zhang, Binyuan Hui, Bo~Zheng, Bowen Yu, Chengyuan Li, Dayiheng Liu, Fei Huang, Haoran Wei, Huan Lin, Jian Yang, Jianhong Tu, Jianwei Zhang, Jianxin Yang, Jiaxi Yang, Jingren Zhou, Junyang Lin, Kai Dang, Keming Lu, Keqin Bao, Kexin Yang, Le~Yu, Mei Li, Mingfeng Xue, Pei Zhang, Qin Zhu, Rui Men, Runji Lin, Tianhao Li, Tianyi Tang, Tingyu Xia, Xingzhang Ren, Xuancheng Ren, Yang Fan, Yang Su, Yichang Zhang, Yu~Wan, Yuqiong Liu, Zeyu Cui, Zhenru Zhang, and Zihan Qiu. 2024.
\newblock Qwen2.5 Technical Report.
\newblock \emph{arXiv preprint arXiv:2412.15115}.

\bibitem[{Ye et~al.(2025)Ye, Zhao, Zhang, and Jiang}]{ye2025unide}
Guanghui Ye, Huan Zhao, Zixing Zhang, and Zhihua Jiang. 2025.
\newblock UniDE: A multi-level and low-resource framework for automatic dialogue evaluation via LLM-based data augmentation and multitask learning.
\newblock \emph{Information Processing \& Management}, 62(3):104035.

\bibitem[{Yerukola et~al.(2024)Yerukola, Vaduguru, Fried, and Sap}]{yerukola2024pope}
Akhila Yerukola, Saujas Vaduguru, Daniel Fried, and Maarten Sap. 2024.
\newblock \href {https://doi.org/10.18653/v1/2024.acl-short.26} {Is the Pope Catholic? Yes, the Pope is Catholic. Generative Evaluation of Non-Literal Intent Resolution in {LLM}s}.
\newblock In \emph{Proceedings of the 62nd Annual Meeting of the Association for Computational Linguistics (Volume 2: Short Papers)}, pages 265--275, Bangkok, Thailand. Association for Computational Linguistics.

\bibitem[{Yu et~al.(2021)Yu, Da~San~Martino, Mohtarami, Glass, and Nakov}]{yu2021interpretable}
Seunghak Yu, Giovanni Da~San~Martino, Mitra Mohtarami, James Glass, and Preslav Nakov. 2021.
\newblock Interpretable Propaganda Detection in News Articles.
\newblock In \emph{Proceedings of the International Conference on Recent Advances in Natural Language Processing (RANLP 2021)}, pages 1597--1605.

\bibitem[{Zhang et~al.(2024{\natexlab{a}})Zhang, Wang, Xu, Zhou, Su, Zhao, Li, Chen, and Gao}]{zhang2024mintrec}
Hanlei Zhang, Xin Wang, Hua Xu, Qianrui Zhou, Jianhua Su, Jinyue Zhao, Wenrui Li, Yanting Chen, and Kai Gao. 2024{\natexlab{a}}.
\newblock \href {https://openreview.net/forum?id=nY9nITZQjc} {{MI}ntRec 2.0: A Large-scale Benchmark Dataset for Multimodal Intent Recognition and Out-of-scope Detection in Conversations}.
\newblock In \emph{The Twelfth International Conference on Learning Representations}, pages 1--27.

\bibitem[{Zhang et~al.(2024{\natexlab{b}})Zhang, Huang, Jin, and Lu}]{zhang2024vision}
Jingyi Zhang, Jiaxing Huang, Sheng Jin, and Shijian Lu. 2024{\natexlab{b}}.
\newblock Vision-language models for vision tasks: A survey.
\newblock \emph{IEEE Transactions on Pattern Analysis and Machine Intelligence}.

\bibitem[{Zhang and Ghorbani(2020)}]{zhang2020overview}
Xichen Zhang and Ali~A Ghorbani. 2020.
\newblock An overview of online fake news: Characterization, detection, and discussion.
\newblock \emph{Information Processing \& Management}, 57(2):102025.

\bibitem[{Zhang et~al.(2021)Zhang, Cao, Li, Sheng, Zhong, and Shu}]{zhang2021mining}
Xueyao Zhang, Juan Cao, Xirong Li, Qiang Sheng, Lei Zhong, and Kai Shu. 2021.
\newblock \href {https://doi.org/10.1145/3442381.3450004} {Mining Dual Emotion for Fake News Detection}.
\newblock In \emph{Proceedings of the Web Conference 2021}, pages 3465--3476. Association for Computing Machinery.

\bibitem[{Zhou and Zafarani(2020)}]{zhou2020survey}
Xinyi Zhou and Reza Zafarani. 2020.
\newblock A survey of fake news: Fundamental theories, detection methods, and opportunities.
\newblock \emph{ACM Computing Surveys (CSUR)}, 53(5):1--40.

\bibitem[{Zhou et~al.(2023)Zhou, Zhu, Yerukola, Davidson, Hwang, Swayamdipta, and Sap}]{zhou2023cobra}
Xuhui Zhou, Hao Zhu, Akhila Yerukola, Thomas Davidson, Jena~D Hwang, Swabha Swayamdipta, and Maarten Sap. 2023.
\newblock Cobra frames: Contextual reasoning about effects and harms of offensive statements.
\newblock \emph{arXiv preprint arXiv:2306.01985}.

\bibitem[{Zhu et~al.(2022)Zhu, Sheng, Cao, Li, Wang, and Zhuang}]{zhu2022generalizing}
Yongchun Zhu, Qiang Sheng, Juan Cao, Shuokai Li, Danding Wang, and Fuzhen Zhuang. 2022.
\newblock \href {https://doi.org/10.1145/3477495.3531816} {Generalizing to the Future: Mitigating Entity Bias in Fake News Detection}.
\newblock In \emph{Proceedings of the 45th International ACM SIGIR Conference on Research and Development in Information Retrieval}, pages 2120--2125. Association for Computing Machinery.

\bibitem[{Zhu et~al.(2023)Zhu, Sheng, Cao, Nan, Shu, Wu, Wang, and Zhuang}]{zhu2022memory}
Yongchun Zhu, Qiang Sheng, Juan Cao, Qiong Nan, Kai Shu, Minghui Wu, Jindong Wang, and Fuzhen Zhuang. 2023.
\newblock \href {https://doi.org/10.1109/TKDE.2022.3185151} {Memory-Guided Multi-View Multi-Domain Fake News Detection}.
\newblock \emph{IEEE Transactions on Knowledge and Data Engineering}, 35(7):7178--7191.

\end{thebibliography}

\end{document}